%% file: main_arxiv.tex
\documentclass[10pt,twocolumn]{article}

\usepackage[a4paper,text={174mm,247mm},columnsep=12pt,centering]{geometry}

\usepackage[T1]{fontenc}
\usepackage{times}

\usepackage{latexsym,ifthen,rotating,calc,textcase}
\usepackage{amsfonts,amssymb,amsbsy,amsmath,amsthm}
\usepackage{caption}
\usepackage{multicol}
\usepackage{tabularx}
\usepackage[dvipsnames, table]{xcolor}
\usepackage{multirow}
\usepackage{algorithm}
\usepackage{algpseudocode}
\usepackage{graphicx}
\usepackage{color}
\usepackage{colortbl}
\usepackage{makecell}
\usepackage{booktabs}

\usepackage[round,authoryear]{natbib}

\usepackage{authblk}

\usepackage[colorlinks,bookmarksopen,bookmarksnumbered,citecolor=blue,urlcolor=blue,linkcolor=blue]{hyperref}

\definecolor{cYes}{HTML}{d4edda}
\definecolor{cDiff}{HTML}{fff3cd}
\definecolor{cNA}{HTML}{e9ecef}
\definecolor{cHeader}{HTML}{2c3e50}
\definecolor{sectionbg}{RGB}{200,216,238}
\definecolor{racingbg}{RGB}{253,232,200}
\definecolor{urbanbg}{RGB}{212,237,218}
\definecolor{simbg}{RGB}{226,217,243}
\definecolor{gripbg}{RGB}{248,215,218}
\definecolor{dtaFull}{HTML}{1F6FAD}
\definecolor{dtaNoIntent}{HTML}{E07B30}
\definecolor{dtaTie}{HTML}{9A9A9A}

\newcommand{\jon}[1]{}
\newcommand{\deepak}[1]{}
\newcommand{\guy}[1]{}
\newcommand{\thomas}[1]{}
\newcommand{\emily}[1]{}
\newcommand{\andrew}[1]{}
\newcommand{\pato}[1]{}
\newcommand{\blue}[1]{#1}
\newcommand{\elliot}[1]{}

\newcommand{\tagRacing}{\colorbox{racingbg}{\footnotesize\textbf{Racing}}}
\newcommand{\tagUrban}{\colorbox{urbanbg}{\footnotesize\textbf{Urban}}}
\newcommand{\tagSim}{\colorbox{simbg}{\footnotesize\textbf{Sim}}}
\newcommand{\tagGRIP}{\colorbox{gripbg}{\footnotesize\textbf{GRIP}}}

\newcolumntype{P}[1]{>{\raggedright\arraybackslash}p{#1}}

\newcommand{\orcid}[1]{\href{https://orcid.org/#1}{\textcolor{green!60!black}{\textbf{iD}}}}

\input{my_commands}

\providecommand{\runninghead}[1]{}
\providecommand{\corrauth}[1]{}
\providecommand{\email}[1]{}
\newcommand{\backmattersection}[1]{\subsection*{#1}}

\newenvironment{funding}{\backmattersection{Funding}}{}
\newenvironment{dci}{\backmattersection{Declaration of conflicting interests}}{}
\newenvironment{sm}{\backmattersection{Supplemental material}}{}

\allowdisplaybreaks

\title{\Large\bfseries Dream to Assist: Intent-Conditioned World Models for Shared Vehicle Control}

\author[1]{Jonathan DeCastro}
\author[1]{Thomas Balch}
\author[1]{Emily Sumner}
\author[1]{Deepak Gopinath}
\author[1]{Laporsha Dees}
\author[1]{Patricio Reyes Gomez}
\author[2]{Elliot Weiss}
\author[1]{Megan Applegate-Kenton}
\author[1]{Todd Rowell}
\author[3]{John Leonard}
\author[2]{Avinash Balachandran}
\author[1]{Guy Rosman}
\affil[1]{Toyota Research Institute, Cambridge, MA 02139, USA}
\affil[2]{Toyota Research Institute, Los Altos, CA 94022, USA}
\affil[3]{Massachusetts Institute of Technology, Cambridge, MA 02139, USA}
\date{}

\begin{document}

\makeatletter
\twocolumn[
  \begin{@twocolumnfalse}
    \maketitle
    \vspace{-1.5em}
    \begin{center}
      \begin{minipage}{0.9\textwidth}
        \small
        \noindent\textbf{Abstract.}\enspace
        Assistive agents in shared-control settings face a dual challenge: acting proactively to align with the human's intentions while remaining legible enough for the human to comprehend and fluently respond in real time.
        We present \dreamtoassist{}, a model-based reinforcement learning framework for tightly-coupled shared decision making that addresses both sides of this challenge. Our approach extends a recurrent state-space world model with intent inference, enabling an assistive policy trained via fictitious co-play to anticipate and support discrete human objectives without overriding human agency. Beyond the computational framework, we describe the full deployment pipeline needed to translate a world-model-based policy into a safe, human-centered system in a resource-efficient way. Crucially, we address the sim-to-real-human gap via policy down-selection and reward restructuring within an iterative human-machine interface design. We validate the framework through three human-subjects studies and a preliminary on-vehicle deployment spanning two qualitatively distinct domains: high-speed racing ($N=30$) and urban driving with richer intent structure ($N=10$). Our results demonstrate that intent conditioning yields measurable improvements in torque alignment and collision rate. We surface key design insights and practices for deploying learning-based assistive agents in safety-critical, real-time human-robot teaming.

      \end{minipage}
    \end{center}
    \vspace{1.5em}
  \end{@twocolumnfalse}
]
\makeatother

{\renewcommand\thefootnote{}\footnotetext{Corresponding author: Jonathan DeCastro, Toyota Research Institute, Cambridge, MA 02139, USA. Email: \href{mailto:jonathan.decastro@tri.global}{jonathan.decastro@tri.global}}\addtocounter{footnote}{-1}}

\input{1_introduction}

\input{2_background}

\input{3_technical_approach}

\input{4_interface_selection}

\input{6_human_evaluation}

\input{8_discussion}
\input{7_conclusions}


\section*{Ethical Considerations}
\subsubsection*{Ethical approval}
This research was conducted under the oversight of WCG IRB on behalf
of Toyota Research Institute (IRB Tracking Number: 20232718,
Study Number: 1356002).

\subsubsection*{Informed consent}
All participants provided written informed consent prior to
participation, covering inclusion in the study, collection and
use of data, and publication of anonymized results.

\begin{dci}
The author(s) declared no potential conflicts of interest with
respect to the research, authorship, and/or publication of
this article.
\end{dci}

\begin{funding}
The author(s) disclosed receipt of the following financial support
for the research, authorship, and/or publication of this article:
This work was supported by the Toyota Research Institute.
\end{funding}

\begin{sm}
Supplementary videos accompany this article:
\begin{itemize}
  \item \textbf{Video~1}: Montage of representative trials from
    Evaluation~1 (Racing Characterization).
  \item \textbf{Video~2}: Montage of representative trials from
    Evaluation~3 (Urban Controlled Validation), showing the
    \textbf{D2A-Full} condition.
  \item \textbf{Video~3}: Demonstration of the GRIP physical
    platform deployment.
\end{itemize}
\end{sm}

\bibliographystyle{plainnat}
\bibliography{references}

\appendix
\input{A_model_architecture}
\input{B_interface_implementation}

\input{C_detailed_methods}
\input{D_additional_results}

\end{document}

%% file: my_commands.tex
\usepackage{xcolor}
\usepackage{url}
\usepackage{amsfonts}
\usepackage{nicefrac}
\usepackage{microtype}
\usepackage{amsmath}
\usepackage{amssymb}
\usepackage{mathtools}
\usepackage{subcaption}
\usepackage{ragged2e}
\usepackage{etoolbox}
\usepackage{xspace}
\usepackage{xpatch}
\usepackage{enumitem}
\usepackage[hang,flushmargin,bottom]{footmisc}
\usepackage{pifont}
\usepackage{listings}
\usepackage{environ}
\usepackage{titlesec}
\usepackage[colorlinks,bookmarksopen,bookmarksnumbered,
  citecolor=blue,urlcolor=blue]{hyperref}
\usepackage[capitalise,noabbrev,nameinlink]{cleveref}
\usepackage{dsfont}
\usepackage{placeins}

\graphicspath{{figures/}}
\renewcommand{\arraystretch}{1}

\makeatletter
\def\hlinewd#1{%
\noalign{\ifnum0=`}\fi\hrule \@height #1 \futurelet
\reserved@a\@xhline}
\makeatother
\NewEnviron{mytabular}[1]{%
  \setlength\heavyrulewidth{1.2pt}
  \setlength\lightrulewidth{.5pt}
  \setlength\aboverulesep{.4ex}%
  \setlength\belowrulesep{.8ex}%
  \begin{booktabs}[expand=\BODY]{#1}
    \BODY
  \end{booktabs}
}

\makeatletter
\DeclareDocumentCommand\todo{g}{%
\def\@message{\IfNoValueTF{#1}{TODO}{TODO: #1}}
\textbf{\textcolor[HTML]{FF8811}{\@message}}
\@latex@warning{\@message}{}{}}
\makeatother

\makeatletter
\newcommand{\removeParBefore}{\ifvmode\vspace*{-\baselineskip}\setlength{\parskip}{0ex}\fi}
\newcommand{\removeParAfter}{\@ifnextchar\par\@gobble\relax}
\newcommand{\eq}{\begingroup\removeParBefore\endlinechar=32 \eqinner}
\newcommand{\eqinner}[2][aligned]{\endlinechar=32%
\begin{gather}\begin{#1}#2\end{#1}\end{gather}\endgroup\removeParAfter}
\makeatother

\DeclareDocumentCommand{\p}{ D<>{p} D<>{} r() }{
\def\content{#3}\patchcmd{\content}{|}{\;#2\vert\;}{}{}
\ensuremath{#1 #2(\content #2)}}

\DeclareDocumentCommand{\P}{ D<>{P} D<>{\big} r() }{
\def\content{#3}\patchcmd{\content}{|}{\;#2\vert\;}{}{}
\ensuremath{\operatorname{#1}#2(\content #2)}}

\DeclareDocumentCommand{\E}{ D<>{E} E{_}{{}} D<>{\big} r[] }{
\def\content{#4}\patchcmd{\content}{|}{\;#3\vert\;}{}{}
\ensuremath{\operatorname{#1}_{#2}#3[\content #3]}}

\DeclareDocumentCommand{\E}{ D<>{E} E{_}{{}} D<>{\big} r[] }{
\def\content{#4}\patchcmd{\content}{|}{\;#3\vert\;}{}{}
\ensuremath{\operatorname{#1}_{#2}#3[\content #3]}}

\DeclareDocumentCommand{\D}{ D<>{D} D<>{\big} r[] }{
\def\content{#3}\patchcmd{\content}{||}{\;#2\|\;}{}{}
\ensuremath{\operatorname{#1}\!#2[\content #2]}}

\NewDocumentCommand{\Nor}{ r() }{\P<Normal>](#1)}
\NewDocumentCommand{\Cat}{ r() }{\P<Cat>](#1)}
\NewDocumentCommand{\Bin}{ r() }{\P<Bin>](#1)}
\NewDocumentCommand{\Bet}{ r() }{\P<Beta>](#1)}
\NewDocumentCommand{\Ber}{ r() }{\P<Bernoulli>(#1)}
\NewDocumentCommand{\Dir}{ r() }{\P<Dir>(#1)}

\DeclareDocumentCommand{\KL}{ D<>{\big} r[] }{\D<KL><#1>[#2]}
\DeclareDocumentCommand{\H}{ D<>{\big} r[] }{\E<H><#1>[#2]}
\DeclareDocumentCommand{\I}{ D<>{\big} r[] }{\E<I><#1>[#2]}

\newcommand{\cmark}{\textcolor{ForestGreen}{\ding{51}}}
\newcommand{\xmark}{\textcolor{red}{\ding{55}}}

\DeclareDocumentCommand{\lnpp}{ D<>{} r() }{
\ensuremath{\p<\ln p_\phi><#1>(#2)}}
\DeclareDocumentCommand{\pp}{ D<>{} r() }{
\ensuremath{\p<p_\phi><#1>(#2)}}
\DeclareDocumentCommand{\ppi}{ D<>{} r() }{
\ensuremath{\p<p^{(i)}_\phi><#1>(#2)}}
\DeclareDocumentCommand{\qp}{ D<>{} r() }{
\ensuremath{\p<q_\phi><#1>(#2)}}
\DeclareDocumentCommand{\qpi}{ D<>{} r() }{
\ensuremath{\p<q^{(i)}_\phi><#1>(#2)}}

\DeclareDocumentCommand{\SymLogNormal}{ D<>{} r() }{
\ensuremath{\p<\operatorname{SymLogNormal}><#1>(#2)}}
\newcommand{\sign}{\operatorname{sign}}

\newcommand{\indc}{\mathds{1}}

\newcommand{\eg}{e.g.\@\xspace}

\newcommand{\dreamtoassist}{\textsc{Dream2Assist}}
\newcommand{\withintentorig}{\textcolor{Magenta}{\textbf{\textsc{D2A}}}}
\newcommand{\initialpolicy}{\textcolor{Magenta}{\textbf{\textsc{D2A}}}}
\newcommand{\refinedpolicy}{\textcolor{ForestGreen}{\textbf{\textsc{D2A}-Refined}}}
\newcommand{\nointent}{\textcolor{Mahogany}{\textbf{\textsc{D2A}-NoIntent}}}
\newcommand{\withintent}{\textcolor{Blue}{\textbf{\textsc{D2A}-Full}}}
\newcommand{\nopolicy}{{\textbf{No-policy}}}

%% file: 1_introduction.tex
\section{Introduction}

In shared control contexts where a human and AI assistant share the same embodiment, assistive agents face a fundamental challenge: to act proactively while remaining aligned with human intent, the agent must anticipate what the human is trying to accomplish before it intervenes.
If it does so in ways that conflict with the human’s intentions, assistance can be perceived as interference rather than support. An effective embodied assistant needs to reason both about the task itself (\eg the end-goal or overall performance of the team) as well as the human users' objectives while performing the task.

This problem is even more apparent in safety-critical, real-time domains, where the human-AI team has to be aligned~\citep{hadfield2016cooperative} and fluent~\citep{hoffman2019evaluating}, the AI has to be understood by the human~\citep{sakai2022explainable}, and the system must avoid risky events \citep{jamieson2022b737}, performance degradation \citep{endsley2023supporting}, or loss of user confidence~\citep{Abbink2018-ei}.
Meeting these requirements simultaneously demands tight integration of policy design, interface design, and human evaluation, and probing how algorithmic choices affect the human's experience of the system as a collaborative partner.

In this work, we present \dreamtoassist{}, a model-based reinforcement learning (RL) framework for real-time shared control. The core idea is to extend a recurrent state-space world model (RSSM) with an intent predictor that infers the human partner's discrete objectives from observed behavior, and to condition the assistive policy's reward on the inferred intent. We instantiate this framework in the domain of shared vehicle control, where the driver and the AI actuate the same steering, throttle, and brake controls.  The result is a policy that is capable of steering with the human's intent in mind rather than merely completing the task, 
which we quantify via torque alignment.

Because real human partners are expensive and variable, such policies are typically trained against simulated human models in simulation; however deploying them with real humans exposes a domain gap between synthetic and real partner behavior and requires more than algorithmic refinement. We describe the full, data-driven deployment pipeline for bridging assistive agents trained in simulation and the target assistive embodiment. In particular, we show the considerations, criteria, and processes needed to fuse robotic machine learning (ML) approaches and human-centric work in order to provide a balance of system performance, safety, and user satisfaction in time-constrained, tightly coupled, human-robot interaction (HRI).

In an initial version of this work~\citep{decastro2025dreaming}, we have shown that a value-aligned assistive policy with intent-conditioned recurrent state space model (RSSM) improves alignment in synthetic settings. This paper further extends the framework to real human partners: we extend our results to allow evaluation with 40+ human participants across three formal studies and more than 30 informal pilots, addressing human factors considerations such as explainability and cognitive load in providing the right level of assistance to the driver.

We validate \dreamtoassist{} across two qualitatively distinct deployment contexts, probing how intent-recognition embedded into such systems' fictitious play with the human enables better performance from real humans, leading to improved overtake success on high-speed straightaways,  %
greater torque alignment with driver intent, and lower collision rates.
A preliminary deployment on a physical vehicle platform ($N = 3$) demonstrates policy transfer with no retraining.

In summary, we make the following contributions: 
\begin{itemize}
    \item We extend \dreamtoassist{}, presented in~\cite{decastro2025dreaming}, with human-data augmentation, reward restructuring, and safety filtering, enabling deployment with real human partners in real-time, safety-critical settings.
    \item We propose a structured deployment pipeline for translating learned interactive policies from simulation to human-centered systems, combining automated policy screening, counterfactual log-replay evaluation, and human-in-the-loop vetting with a coupled iterative interface-design process. We formulate key performance indicators for physical HRI in common multi-agent driving scenarios.
    \item We validate \dreamtoassist{} across two domains via human subject evaluations in high-speed racing ($N_1=24$, $N_2=6$) and surprise-rich urban scenarios ($N=10$), demonstrating that intent conditioning produces measurable improvements over the no-intent-conditioning baseline in torque alignment and collision avoidance under double-blind conditions.
    \item We explore early findings of our approach in a real test vehicle ($N=3$).
\end{itemize}

%% file: 2_background.tex
\section{Related Work}
The approach we present lies at the intersection of several fields, namely ML and RL for HRI, shared control approaches in driving, and interface approaches for time-critical human-machine interactions. While we focus on shared-embodiment driving, our framework is equally applicable to turn-taking HRI settings.

\subsection{Machine Learning Approaches for HRI}

Significant work has been done on sharing control \citep{Abbink2018-ei} and human-robot interaction \citep{sheridan2016human,Onnasch2021-nl}, exploring topics ranging from ergonomics and physiology to language-based strategic teaming.
Prior research has presented learning-based approaches~\citep{semeraro2023hrc} to human-robot interactions. These works focus on a range of topics, including joint-representation learning \citep{Xie2021-id,Reddy2018-ta,Tucker2022-na,Karamcheti2022-ut, bobu2023sirl,sucholutsky2023getting,parekh2023learning,Jeon2020-ki}, data-efficiency \citep{strouse2021collaborating}, intent or plan modeling \citep{Kelley2013-gq,Thill2017-ut,losey2018review,Cheng2020-rf,rudenko2020human,hoffman2024inferring}, hypothesis-space specification and explanation \citep{sucholutsky2023getting,9007490,setchi2020explainable}, alignment~\citep{hoffman2019evaluating,dragan2013legibility}, or other cognitive and social motivations \citep{lebiere2013cognitive,mutlu2016cognitive,admoni2017social}. Other works focused on specific approaches for planning and control, including imitation learning~\citep{cardenas2024xbg,shi2025review}, model-predictive planning~\citep{Schwarting2017-fh,zhang2025nmpc}, or reinforcement learning~\citep{Reddy2018-ta,akalin2021reinforcement, xiao2026implementation}. We go beyond prior work by embedding intent inference directly within the world model's latent dynamics, conditioning the assistive policy's reward on the inferred intent. We further evaluate the full pipeline with human partners in real time.

\subsection{Planning for Shared Control in Driving}

In the context of driving, shared control has been explored for planning approaches~\citep{bobier2010sliding,Bobier2013-uf,10202565,Schwarting2017-fh, benloucif2019cooperative,liang2021shared,dann2021multi,jiang2021goal,macke2021expected} and learning-based approaches \citep{Griffiths2005-md,Mulder2012-fa,Balachandran2016-sl,Broad-RSS-17,Ghasemi2018-dl}. %
Prior work has also considered incorporating game-theory into shared control  \citep{Ghasemi2018-dl,Jugade2019-qg}, as well as explicit human-centric design considerations  \citep{Deng2022-cy, Xie2022-jb}. We refer readers to \citep{Marcano2020-jw, Sarabia2023-vy} for comprehensive reviews. Beyond shared control, significant literature explored intent prediction in driving, see \citep{Lefevre2014-aj,Rudenko2020-cp}, and references therein.

Recent work has proposed a shared-control model using model-predictive control that leverages predicted trajectory information \citep{10202565}, therefore implicitly capturing driver intent. Additional recent work has augmented a model-predictive controller with the ability to explicitly infer driver intent \citep{Weiss2023-ti}, enabling the controller to share the steering, throttle, and brake actions with a human. In our approach, we capture discrete driver intent in a way that is conducive to semantically-meaningful, multi-modal continuous behaviors (e.g., going left \textit{or} right around an obstacle). Further, by framing our task as a multi-agent reinforcement-learning problem, our proposed solution can extend to multiple agents planning~\citep{busoniu2008comprehensive,huh2023multi} much more naturally than prior work. 
Finally, we frame our decision-making approach within a recurrent state-space model \citep{hafner2019dream} which is extended to infer the human's objective, building on recent work that has explored hierarchical or hybrid state abstractions \citep{Zholus2022-wt,Gumbsch_undated-tr}. 

\subsection{HRI and Interfaces in Time-Critical Settings}

Deploying assistive interactions in real-time safety-critical settings includes challenges beyond policy design -- accounting for human behaviors within the interaction due to latency~\citep{mackenzie1993lag,mcruer1974mathematical,Xu2017-vn,rieger2022human,semeraro2023hrc}.

Challenges in design often include mode confusion and awareness under time stress, first contact usability with short exposure, HMI signal design for rapid perception, as well as experimental design validity in such conditions~\citep{albers2020usability,dong2024review}. 
Situation awareness frameworks~\citep{endsley1995toward} have been widely applied to understand failures in human-automation teaming under time pressure and have been adapted for driving~\citep{walker2008feedback} and to HRI more broadly~\citep{van2024agent}.
The timing and modality of feedback signals are particularly consequential in high-speed settings, with haptic feedback shown to convey urgent cues with lower cognitive overhead~\citep{ploch2017comparing}.
Transparency and explainability of AI actions present an additional design dimension~\citep{sakai2022explainable,setchi2020explainable}, with trajectories often used to convey robotic intent for shared control interactions~\citep{brooks2020visualization}.

While these aspects have been covered in existing literature, the integration of learning-based time-critical human-in-the-loop systems expands the design space, while reducing our ability to control the concepts the system exposes to the user, requiring careful probing of the common assumptions on these aspects. %

\subsection{Design Metaphors for Shared Control}

The behavior of a partially automated vehicle is intelligible to the driver only if the driver has formed an accurate mental model of how it works~\citep{norman2013design}; design metaphors are a long-standing tool for inducing such a model by grounding novel functions in familiar imagery. The most influential metaphor in shared driving control is Flemisch et al.'s H-Metaphor~\citep{flemisch2003h}, which casts the driver–vehicle relationship as horse-and-rider, with continuously variable ``loose-rein'' and ``tight-rein'' coupling enacted through bidirectional haptic communication. The H-Metaphor was subsequently realized as a concrete haptic-multimodal interaction architecture (H-mode) for cooperative guidance and control~\citep{altendorf2015hmode}, and generalized by Flemisch and colleagues into a unifying framework spanning joint action, cooperative guidance, and shared/traded/supervisory control~\citep{flemisch2019joining}. 

Adjacent metaphors recast the same problem from different perspectives: van Paassen et al.\ enumerate the design choices of haptic shared control through the lens of an instructor teaching a student to fly~\citep{vanpaassen2017four}. Ju proposes that many driver–vehicle directives are best modelled as implicit interactions of the kind a passenger has with the driver~\citep{ju2015design}; and Zhang et al.\ argue that drivers should be framed as the commander of a crewed vehicle rather than as a passive fallback~\citep{zhang2021fallback}. %
\dreamtoassist{} may be characterized by a modified commander-crew metaphor, where the driver (a ``hands-on'' captain) leads through action, and the assistive policy must infer the driver's high-level intent from observed behavior and resolve it into assistive continuous controls.  It can override when the inferred intent would lead to an unsafe outcome.

%% file: 3_technical_approach.tex
\section{Technical Approach}
\label{sec:approach}

\subsection{Preliminaries}
We frame the model-based reinforcement learning (MBRL) problem as a two-player (human ego driver and assistive agent ego driver) partially-observable Markov decision process (POMDP) defined by the tuple $M = \langle\mathcal{X}^{\kappa}, \mathcal{O}^{\kappa}, \mathcal{A}^{\kappa}, \mathcal{T}^{\kappa}, \mathcal{R}^{\kappa}, \gamma\rangle^{\kappa=H, A}$, where, for agent $\kappa\in \{H, A\}$ (for the human and AI agent, respectively), $\mathcal{X}^{\kappa}$ denotes the space of latent world states spaces, $\mathcal{O}^{\kappa}$ denotes the observation space with observations generated according to an emission function $e^{\kappa}: \mathcal{X}^{\kappa} \mapsto \mathcal{O}^{\kappa}$, $\mathcal{A}^{\kappa}$ denotes the agent's (continuous or discrete) actions, $\mathcal{T}^{\kappa}: \mathcal{X}^{\kappa} \times \mathcal{A}^{\kappa} \mapsto [0, 1]$ is the transition probability, $\mathcal{R}^{\kappa}: \mathcal{X}^{\kappa} \times \mathcal{A}^{\kappa}  \times \mathcal{X}^{\kappa} \mapsto \mathbb{R}$ is a reward function, and $\gamma \in [0, 1]$ is a discount factor. We aim to train both agents such that they maximize their expected returns $R^{\kappa} = \mathbb{E} \left[ \sum_{t = 1}^T r^{\kappa}_t \right]$.  

Crucially, in the semi-cooperative shared control setting, each reward $r^\kappa_t$ is factored into sub-components, with both sharing the same task (driving) rewards, but where $r^H_t$ contains an additional term for a human's objective, and $r^A_t$ contains additional terms to weaken its contribution in relation to the human's and enforces alignment to the human.

We assume access to a set of task rewards $r_t^{task}$, and a set of policies that capture human behaviors $\Pi_i^H$ for each intent $i \in \mathcal{Y}$.  Here, \emph{task} refers to the domain-level objective shared
by both agents (e.g., completing the race, navigating the town)~\citep{michon1985critical}, while \emph{intent} refers to the human's local, moment-to-moment objective within that task (e.g., passing the car ahead vs.\ staying behind it). For each intent (e.g. passing, changing lanes, stopping, etc.), we then assume a label of human intent, or local objectives, that the human could be deciding at any given time point.

The intent space $\mathcal{Y}$ is domain-dependent; in this work, we consider the intent space to be discrete to match different maneuvers that could be taken. %
In racing, $\mathcal{Y} = \{\text{pass}, \text{stay behind}\}$: the driver either intends to overtake the lead vehicle or remain behind it. In urban driving, $\mathcal{Y}$ combines lateral and longitudinal components: $\mathcal{Y} = \{\text{left}, \text{right}\} \times \{\text{go}, \text{slow}, \text{stop}\}$, capturing both the direction of avoidance and the desired speed response to a hazard. 
The reward terms conditioned on intent are defined in Secs.~\ref{sec:pop_human_model}--\ref{sec:assistant_objectives} and detailed per domain in Appendix~\ref{sec:policy_interface_design}.

\subsection{\dreamtoassist{}}
\label{subsec:comp_dream2assist}
The computational backbone of  \dreamtoassist{} underlying our work
learns a common latent representation in the structure of a recurrent state space model (RSSM)~\citep{Hafner2023-rz, hafner2019learning} to build a hybrid discrete/continuous state of the human partner's behavior, their rewards, and intent for achieving certain goals. RSSMs have been shown to be suitable for many domains (e.g. locomotion, Atari, Minecraft).  We extend the usual RSSM formulation to jointly represent behavior and intent, conditioning the assistant's reward on the inferred human objective.
To accommodate both a \textit{human} and \textit{assistive} agent operating in a collaborative setting, we train the assistive agent alongside human agents, with both agents sharing actions taken in an environment. Each agent receives a reward for each action taken, and each are trained to (1) build an accurate RSSM and (2) learn how to act in order to maximize its own expected returns~\citep{sutton2018reinforcement}. One challenge in collaboration is that part of the assistive agent's world model includes aspects of the future human's plan, such as preferences, desires, and goals of the human (these are often collectively referred to as objectives~\citep{9007490} or intents~\citep{losey2018review} in the literature, and we will use these terms interchangeably in the paper), which are often unknown and only weakly observed by their behavior~\blue{\citep{9007490}}. We extend the RSSM formulation to allow it to be supervised on a diverse set of humans, whose objectives are known and labeled, to force the representation to be jointly aware of the behavior and intent across a variety of human types and physical environments. We then devise a scheme that feeds in inferred human intent into the assistant's reward function.

\begin{figure*}
    \centering
    \includegraphics[trim={0 3.0cm 0 3.0cm}, clip, width=0.9\linewidth]{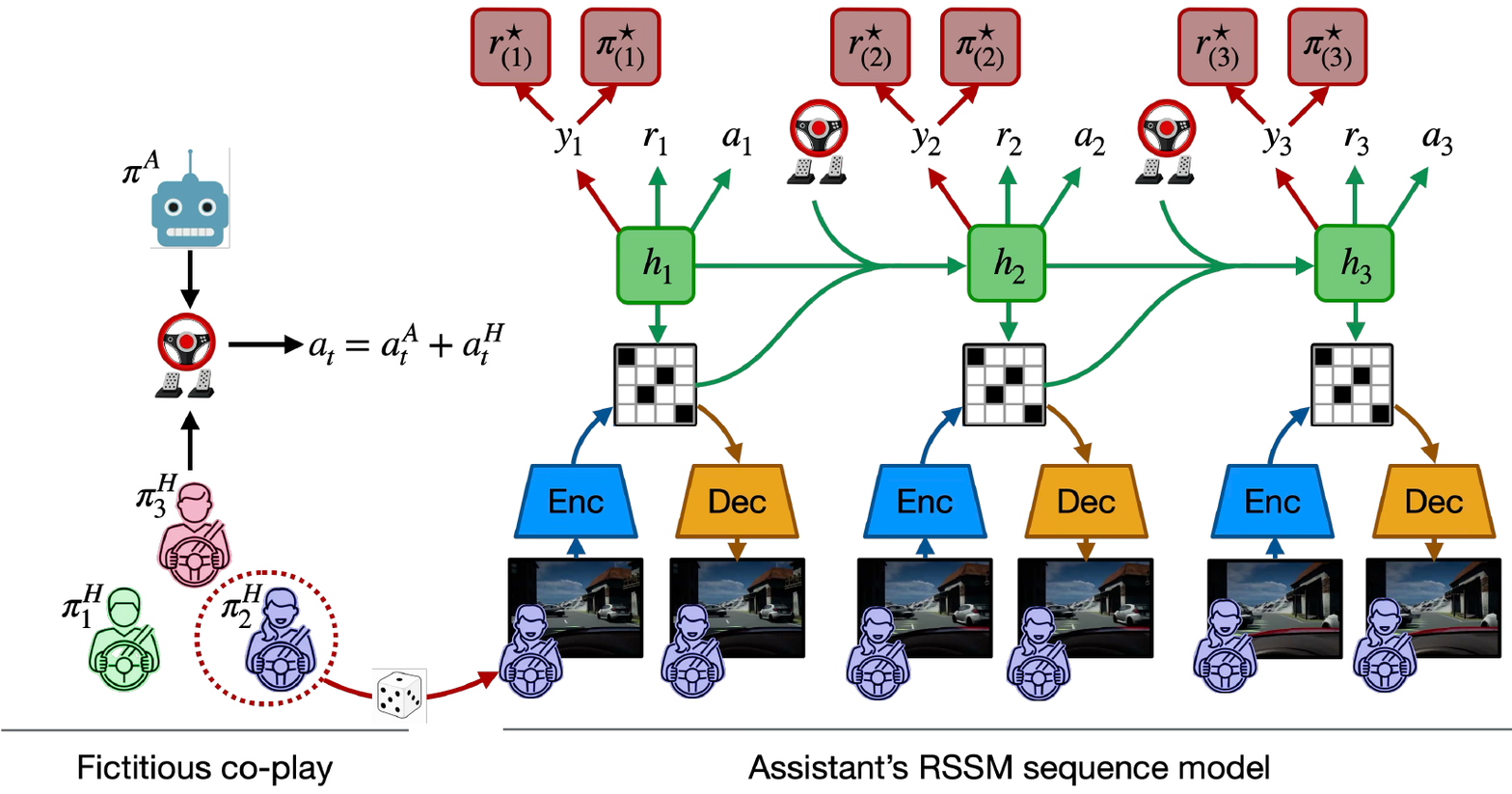}
    \caption{\dreamtoassist{} RSSM:  We blend human actions linearly with the actions from an unfrozen assistive agent. For both human and assistive agents, a multi-head RSSM architecture is then used to predict the observations, reward, and human intent (assistive only), trained to maximize their log-likelihoods against samples taken from the environment. The intent head is trained to match frozen human intents, which are fixed a priori. The assistive agent's rewards are shaped based on the \textit{optimal policy} $\pi^*_{\hat{y}^A}$ and optimal predicted reward $r^*_{\hat{y}^A}$ for inferred intent $\hat{y}^A$.}
    \label{fig:overview}
\end{figure*}

In order to allow the system to maintain an ability to reason over the preferences of the human independent of that of the joint human-robot system, our approach considers both the human driver and the assistive agent planner as separate models. The training process is outlined in Fig.~\ref{fig:overview}.

In the cooperative setting, the model of the environment follows a structure in which there are certain task-specific rewards which are available to both human and assistive agents, with the key distinctions being human-objective (specific to the human), and intervention penalties (specific to the assistant).

\paragraph{Task-Specific Rewards} 
We decompose the task reward into components shared across domains:
\begin{equation}
\label{eq:task_specific_rews}
        r^{task}_t = r^{collision}_t + r^{bounds}_t + r^{progress}_t
\end{equation}

where $r^{collision}_t$ penalizes proximity to other vehicles, $r^{bounds}_t$ penalizes departures from the road, and $r^{progress}_t$ rewards progress toward the domain-specific goal (e.g. reaching the finish line in racing, or safely steering around an wrong-way car in urban driving). The precise definition of each reward term is detailed in Appendix~\ref{sec:policy_interface_design}.

\paragraph{Intent-Aware World Model}
The objective of the world model is to provide a representation that the agent can use to interact with the driver and the world, and we posit that this representation can support the agent to reason jointly about both the task and human objectives. We build off of the RSSM of DreamerV2/V3~\citep{hafner2020mastering, Hafner2023-rz}, according to the architecture in Figure~\ref{fig:overview}. For agent $\kappa\in \{H, A\}$ (respectively, the \textbf{H}uman and \textbf{A}ssistive agent), the RSSM model, parameterized by $\phi$ and denoted $W_\phi^{\kappa}$, includes:
\eq{
\begin{alignedat}{4}
& \text{Discrete rep. encoder $\boldsymbol{z}^\kappa_t$:}   && \boldsymbol{z}^{\kappa}_t            &\ \sim &\ q_\phi^{\kappa}(\boldsymbol{z}^{\kappa}_t | \boldsymbol{h}^{\kappa}_t, \boldsymbol{x}_t) \\
& \text{Recurrent state sequence $\boldsymbol{h}^\kappa_t$: }       && \boldsymbol{h}^{\kappa}_t           &\ =    &\ f_\phi^{\kappa}(\boldsymbol{h}^{\kappa}_{t-1}, \boldsymbol{z}^{\kappa}_{t-1}, \boldsymbol{a}^{\kappa}_{t-1}) \\
& \text{Dynamics predictor:}   && \hat{\boldsymbol{z}}^{\kappa}_t      &\ \sim &\ p_\phi^{\kappa}(\hat{\boldsymbol{z}}^{\kappa}_t | \boldsymbol{h}^{\kappa}_t) \\
\end{alignedat}
}

where $\boldsymbol{x}_t$ denotes the input observation. Following DreamerV3, the latent state uses a discrete categorical representation that is independent of the intent space $\mathcal{Y}$.  The output heads are similar to the DreamerV2/V3 architecture, with the addition of an intent predictor for the assistant agent (in {\color{BrickRed}red}), and are all bottlenecked on the hidden states $\boldsymbol{s}_t^{\kappa} = \{\boldsymbol{h}_t^{\kappa}, \boldsymbol{z}_t^{\kappa}\}$,
\eq{
\begin{alignedat}{8}
\label{eq:preds}
& \text{Decoder:}\       && \hat{\boldsymbol{x}}^{\kappa}_t      &\ \sim &\ p_\phi^{\kappa}(\hat{\boldsymbol{x}}^{\kappa}_t | \boldsymbol{s}^{\kappa}_t)
& \quad \text{Continue:}     && \hat{c}^{\kappa}_t      &\ \sim &\ p_\phi^{\kappa}(\hat{c}^{\kappa}_t | \boldsymbol{s}^{\kappa}_t) \\
& \text{Reward:}       && \hat{r}^{\kappa}_t      &\ \sim &\ p_\phi^{\kappa}(\hat{r}^{\kappa}_t | \boldsymbol{s}^{\kappa}_t)
& \quad \color{BrickRed}\text{Intent:}\        && \color{BrickRed}\hat{y}^A_t      &\ \color{BrickRed}\sim &\ \color{BrickRed}p_\phi^A(\hat{y}^A_t | \boldsymbol{s}^A_t) \\
\end{alignedat}
}
The AI intent head predicts the intent of the human agent, and we assume at least partial access to the intent labels for a given episode, which takes the form of an integer-valued target function.  We normalize the output by predicting the symlog ($\sign(y) \ln(\vert y \vert + 1)$) of the output, and use a discrete distribution to predict $y$ using the two-hot encoding of~\citep{Hafner2023-rz}. The parameters $\phi$ of the world model $W_\phi^{\kappa}$ are trained to minimize the loss
\begin{align*}
    \mathcal{L}^{\kappa}(\phi) &= \mathbb{E}_{q_{\phi}^{\kappa}}\left[\beta_{pred}\mathcal{L}_{pred}^{\kappa} + \beta_{KL}\mathcal{L}_{KL}^{\kappa}\right]
\end{align*}
The prediction loss $\mathcal{L}_{pred}^{\kappa}$ minimizes the likelihood under the predictor distributions in Eq.~\eqref{eq:preds}, $\mathcal{L}_{KL}^{\kappa}$ minimizes the KL divergence between the prior $p^{\kappa}_{\phi}$ and the approximate posterior $q^{\kappa}_{\phi}$, and $\beta_{pred}$ and $\beta_{KL}$ are scalar weighting values. Given an intent label $y_t$, the portion of loss $\mathcal{L}^A_{pred}$ for the intent predictor is taken as the negative log-likelihood of the label under $p_{\phi}^A$.

The world model training is alternated with training for the behavior model governing actions $\boldsymbol{a}^\kappa_t$, with the behavior model learned via an actor-critic policy training over the estimates $\boldsymbol{s}_t$, $\hat{r}_t$.
\eq{
&\text{Actor:}
&& \boldsymbol{a}^{\kappa}_t \sim \p<\pi^{\kappa}_\theta>(\boldsymbol{a}^{\kappa}_t|{\boldsymbol{s}}^{\kappa}_t)
&& \text{Critic:}
&& v^{\kappa}_\psi(\boldsymbol{s}_t) \approx \mathbb{E}_{p^{\kappa}_\phi,\pi^{\kappa}_\theta}[\hat{r}^{\kappa}_t]
\label{eq:ac}
}
The critic is trained to minimize the temporal-difference loss on the value function $v^{\kappa}_{\psi}$. The actor attempts to maximize the critic-predicted value. For further details, we refer the reader to~\citep{Hafner2023-rz}, which introduces the DreamerV3 architecture that our RSSM builds upon. The primary extensions in our work are the dual-agent formulation (separate human and assistant models sharing a common world model), the intent predictor head \eqref{eq:preds}, and the intent-conditioned reward structure (Appendix~\ref{sec:policy_interface_design}).

\subsection{Population-Based Human Modeling}
\label{sec:pop_human_model}
Similar to recent work describing human preferences for a task using reward shaping~\citep{tang2021discovering, yu2023learning}, we train a population of simulated human policies to serve as training partners for the assistive agent.  Different from these works, we group labeled sub-populations according to certain inherent objectives or preferences. We train several sub-populations using a separate human reward $r^H_t = r^{task}_t + r^{y}_t$, which is composed of the base task rewards in Eq.~\eqref{eq:task_specific_rews}, with the addition of several distinct \textit{human objectives}, spanning different multimodal behaviors of the human. These are then formed into a collection of humans $\{\pi^H_1, \pi^H_2, \ldots, \pi^H_N\} \in \{\Pi^H_1, \Pi^H_2, \ldots, \Pi^H_N\}$ for $N$ different objectives. Examples of different settings for $r^{y}_t$ are given in Appendix~\ref{sec:policy_interface_design}.

Humans are error-prone and pursue their objectives irrationally~\citep{chipman2016oxford}. To capture this, we partially train simulated human partners (hereafter, referred to as \emph{human policies}) to pair with the assistive agent in a process known as fictitious co-play (FCP)~\citep{strouse2021collaborating}, yielding a population of human policies $\{\pi^H_{y, k}\}_k \cup \{\pi^*_y\}$ for each intent sub-population $\Pi^H_y$, where $k$ represents the policy checkpoint for some amount of completed training. Using the FCP framework, each checkpoint $k \in [1, K]$ is an agent trained to $k\%$ completion and the final checkpoint $k=K$ is an optimal policy. Each checkpoint is endowed with a suboptimal RSSM world model, with the final checkpoint ($k = K$) approximating the optimal policy. While we apply FCP in this work, we note that our approach is agnostic to how actual human behaviors are generated and future work may consider other approaches to synthesizing human policies \citep{park2023generative,venuto2024code}. 

\subsection{Alignment with the Human}
To learn alignment with human drivers, we expose the assistive agent to different human policies at training time. Training with human policy partners in a fictitious co-play setting~\citep{strouse2021collaborating} allows the assistant to become robust to different possible human behaviors, but does not teach the assistant to distinguish between the discrete modes of human behavior, which is necessary for assistance in our driving task. 

If trained to optimality, the human policies induce optimal behaviors $\pi^{\star}_y$ for each intent $y$.  We refer to these as \textit{expert policies} for each enumerated objective.
At deployment, the intent predictor estimates $\hat{y}_t^A$ from the latent state, and the assistive reward is shaped to match the expert policy corresponding to the inferred intent.
The human policy $\pi^H_{y_t}$ may adopt an intent $y_t$ that varies with time. To simplify training, we train from drivers and rollouts (simulated episode trajectories) with fixed intents, $y$, though our problem setup generalizes to variable-intent scenarios (e.g., a human switching from ``following'' to ``passing'' during a race).  In practice, the intent predictor re-estimates $\hat{y}_t^A$ at every timestep from a moving window of latent states, so the system can track intent changes that occur in real-time even though individual training episodes use fixed labels.  The training distribution nonetheless covers all intents, exposing the world model to the full range of human behaviors without requiring explicit transition labels during training. Extending the training procedure to explicitly model intent transitions is a direction for future work (Sec.~\ref{sec:discussion}).  When the intent predictor assigns similar probability to multiple intents, the policy defaults to task-completion behavior (minimizing intervention), as the intent-conditioned reward components approximately cancel.

We train a separate Dreamer agent for each intent class. To reduce the domain gap between synthetic and real human control signals and induce more human-like in behavior, we apply a smoothing operation to ensure the human policies provide smooth control inputs~\citep{mysore2021regularizing}.
We additionally mix real human driving data to ground the world model in realistic dynamics by labeling the state-action pairs with rewards through an offline re-simulation process. Since there is no ground-truth intent for the entire trajectory, we do not label these additional trajectories with intent. We keep the proportion small (20\% of the replay buffer, as performance was not sensitive in the proportion range evaluated) in order to reduce the off-policy drift between the human policy and actual human data, and furthermore to keep the intent head from being starved of supervision labels. Lastly, we re-balance intent classes during training of the AI to prevent mode collapse toward a dominant intent.

\subsection{Value-Aligned Assistive Objectives}
\label{sec:assistant_objectives}
The core principle behind training the assistive policy is to (1) produce actions consistent with what an intent-aligned expert would select, given the human's inferred preference, 
and (2) minimize interventions.
More precisely: at each timestep the assistant infers the human's current intent $\hat{y}_t^A$ from the latent state and retrieves two quantities from the corresponding optimal human world model: the reward $\hat{r}^*(\hat{y}_t^A)$ that an intent-aligned expert would receive, and the action $\boldsymbol{a}^*(\hat{y}_t^A)$ that expert would select.
The assistant is rewarded for achieving reward values consistent with what an intent-aligned expert would achieve, while producing joint human-robot actions that resemble what the optimal human policy would select.
Expert alignment and intervention minimization together allows the policy to explicitly act with the human agent, rather than just completing the task.

For continuous actions $\boldsymbol{a}_t^H, \boldsymbol{a}_t^{A} \in \mathbb{R}^m$ for the human and assistant, respectively, we encourage the assistive agent to minimize its action intervention by adopting a reward of the form
\begin{align*}
r^{interv}_t = -\|\boldsymbol{a}_t^{A}\|_2
\end{align*}
which penalizes the magnitude of the intervention according to an $L_2$-norm. Aggregated over time, the intervention cost forms a mixed $L_1 - L_2$ norm  \citep{Elad_undated-wi}, encouraging temporally sparse interventions. 

The agent is further rewarded for task completion and penalized for driving out of bounds and for collisions (see~\cite{chenlearn23}). These sparse negative rewards serve to encourage the assistant to act as a ``racing guardian,'' keeping the driver on the track and avoiding collisions while making progress towards the finish line. Note that these rewards align with the \textit{task} objective %
but do not favor any particular \textit{human} objective or intent.

Finally, the assistive agent receives a reward for aligning with the inferred objective of the \textit{human driver}. First, the agent receives a reward equal to the inferred reward under the optimal human's world model. 
Second, the agent receives reward for maximizing the likelihood of the human--robot combined action under the optimal human policy conditioned on the predicted intent, $\pi^*(\boldsymbol{a}^A_t + \boldsymbol{a}^H_t \mid \boldsymbol{x}_t, \hat{y}^A_t)$.
Note that $\boldsymbol{a}^H_t$ is the current sub-optimal human action. Intuitively, these rewards encourage the assistant to: (1) select actions for which the optimal human assigns high value and (2) select actions that bring the sub-optimal human closer to the optimal human. Under this setup, although we infer $\hat{y}^A_t$ from the joint human-robot trajectory, we consider it to be more aligned with the \textit{human's} intent, due to: (1) the presence of the intervention sparsity term $r^{interv}_t$, and (2) the fact that the inferred expert alignment term tilts the joint behavior toward the inferred expert human's behavior.

The complete reward for the assistant combines three objectives. The \emph{task reward} $r^{task}_t$ rewards progress and penalizes collisions and out-of-bounds driving.  The \emph{intervention cost} $r^{interv}_t = -\|\boldsymbol{a}_t^{A}\|_2$ penalizes the magnitude of the assistant's action; aggregated over time this forms a mixed $L_1$--$L_2$ norm~\citep{Elad_undated-wi}, encouraging temporally sparse interventions.  The \emph{expert-alignment reward} encourages actions consistent with the inferred optimal human policy.  The combined reward is:
\begin{align}
    \label{eq:ai_reward}
    r^A_t &= \underbrace{r^{task}_t + r^{interv}_t}_{\textrm{task + intervention}} \nonumber\\ 
    &+ \underbrace{\alpha_r \hat{r}^*_t(\hat{y}^A_t) + \alpha_a \|\boldsymbol{a}^H_t + \boldsymbol{a}^{A}_t - \boldsymbol{a}^*_t(\hat{y}^A_t)\|_2}_{\textrm{expert alignment}}
\end{align}
where $\hat{r}^*(\hat{y}^A_t)$ is the reward predicted by the optimal human's world model for the current state-action pair and predicted intent, and $\boldsymbol{a}^*(\hat{y}^A_t) \sim \pi^*(\cdot \mid \boldsymbol{x}_t, \hat{y}^A_t)$ is the corresponding optimal action. Scalars $\alpha_r$ and $\alpha_a$ weight the expert-alignment terms; all evaluations in this paper use the reward-only case ($\alpha_r = 1$, $\alpha_a = 0$), though we compare alternative weightings in~\cite{decastro2025dreaming}.

\subsection{Safety Layer}
\label{sec:safety_layer}
Although the assistive policy is trained to encourage safe joint actions, no learned system can guarantee safety in all states~\citep{oh2025safety}.  Following the framework of~\cite{oh2025safety}, we therefore layer a control-barrier-function (CBF) safety filter on top of the learned policy.  As in that work, the filter treats the human as an unmodeled agent whose inputs enter through the dynamics rather than through an explicit human model; the CBF constraint ensures forward invariance of a safe set defined over the ego vehicle's state regardless of the human's future actions.

This separation is particularly important for shared control with a trajectory visualization, which renders the policy's planned path several seconds ahead: safety violations in the predicted trajectory are visible to the driver and can erode trust before they materialize in the vehicle's actual motion.  The safety filter can be applied to the policy's commanded actions, to the predicted trajectory shown to the driver (Appendix~\ref{sec:traj_viz}), or both.  In the human studies reported here, the filter was applied only to the trajectory visualization in order to isolate the effect of intent conditioning on the assistive torque (Sec.~\ref{sec:evaluation}); extending it to the actuation channel is a natural next step.  We describe the specific implementation used in each evaluation in Sec.~\ref{sec:evaluation}.

In practice, the safety filter activates only when the system approaches a constraint boundary; during normal operation, the learned policy's actions pass through unmodified.

We follow a simplified version of driving safety filters~\citep{oh2025safety}.  
In particular, we assume a function $h(\boldsymbol{x}) \geq 0$ to characterize the safety of the vehicle. Given $\dot{h}$, the gradient of the fully-observable dynamics (can be the true system dynamics or RSSM -- in our case, we use the true system dynamics), we aim to satisfy the first order control barrier function condition to guarantee safety:
\begin{equation}
\label{eq:cbf}
    \dot{h}(\boldsymbol{x}, \boldsymbol{a}) + \alpha h(\boldsymbol{x}) \geq 0
\end{equation}
where constant $\alpha > 0$ and $\dot{h}(\boldsymbol{x}, \boldsymbol{a})$ denotes the time derivative of $h(x)$ along the trajectory of the system. 
We are interested in finding the smallest $\delta \boldsymbol{a}$ with $\boldsymbol{a} = \boldsymbol{a}^H + \boldsymbol{a}^A + \delta \boldsymbol{a}$ such that Eq.~\eqref{eq:cbf} holds, i.e.,
\begin{equation}
    \min_{\delta \boldsymbol{a}}\Vert\delta \boldsymbol{a}\Vert \qquad \textrm{s.t.} \qquad L_g h(\boldsymbol{x}) \delta \boldsymbol{a} \geq -\alpha h(\boldsymbol{x}) - \dot{h}(\boldsymbol{x}, \boldsymbol{a})
\end{equation}
where we assume control-affine dynamics $\dot{\boldsymbol{x}} = f(\boldsymbol{x}) + g(\boldsymbol{x})\boldsymbol{a}$, which is standard for kinematic vehicles and renders Eq.~\eqref{eq:cbf} linear in the control input $\boldsymbol{a}$, and $L_g h(\boldsymbol{x}) = \nabla h(\boldsymbol{x})^\top g(\boldsymbol{x})$ denotes the Lie derivative of $h$ along the control vector field $g$.  Since the dynamics of other vehicles can be predicted within a short time horizon, we can solve this problem analytically:
\begin{equation}
    \delta \boldsymbol{a} = \max\left(0, \frac{-\alpha h(\boldsymbol{x}) - \dot{h}(\boldsymbol{x}, \boldsymbol{a})}{L_g h(\boldsymbol{x})}\right)
\end{equation}
Intuitively, this constraint ensures that if the system is currently safe ($h(\boldsymbol{x}) \geq 0$), then it remains safe at the next timestep (the barrier function cannot decrease at a rate faster than $\alpha h(\boldsymbol{x})$). This may be applied to lane boundaries and static and dynamic obstacles.  In practice, the safety filter activates only when the system approaches a constraint boundary; during normal operation, the learned policy's actions pass through unmodified.

\subsection{Alignment Metrics}
\label{sec:alignment_metrics}
We propose the following metrics for evaluating alignment in any shared-control setting.
Let $\tau^A(t)$ be the signed assistive steering torque at time $t$, $y^H(t) \in \{-1, +1\}$ be a binary intent direction, and $T$ the length of the window.  We define three alignment metrics:
\begin{align}
\langle |\tau^A| \rangle &= \frac{1}{T}\sum_{t=1}^T |\tau^A(t)|
\label{eq:torque_magnitude} \\
\Delta\tau %
    &= \frac{1}{T}\sum_{t=1}^T |\tau^A(t)|
       \Big(  \indc[\operatorname{sign}(\tau^A(t)) = y^H(t)] \nonumber \\
    &\phantom{{}= \frac{1}{T}\sum_{t=1}^T |\tau^A(t)|\Big(}
           - \indc[\operatorname{sign}(\tau^A(t)) \neq y^H(t)] \Big)
    \label{eq:torque_difference} \\
N_{\mathrm{align}} &= \sum_{t=1}^T
\indc\left[\operatorname{sign}(\tau^A(t)) = y^H(t)
\ \wedge\ |\tau^A(t)| > \tau_{\min}\right]
\label{eq:alignment_events}
\end{align}
Here, \eqref{eq:torque_magnitude} captures the average strength of the AI's steering intervention, regardless of direction, \eqref{eq:torque_difference} separates the AI's torque into intent-aligned and intent-opposing components (positive values indicate the AI applies more force when aligned with the human's goal than when opposing it), and Eq.~\eqref{eq:alignment_events} counts timesteps where the AI's torque is both directionally consistent with the human's intent and exceeds a minimum threshold $\tau_{\min}$, filtering out negligibly small corrections.  These metrics provide a common evaluation framework across all experiments reported in Sec.~\ref{sec:evaluation}.

%% file: 4_interface_selection.tex
\section{Application Domains \& Test Platforms
}
\label{sec:domains}
\begin{figure}[tb]
    \centering
    \includegraphics[width=0.99\linewidth]{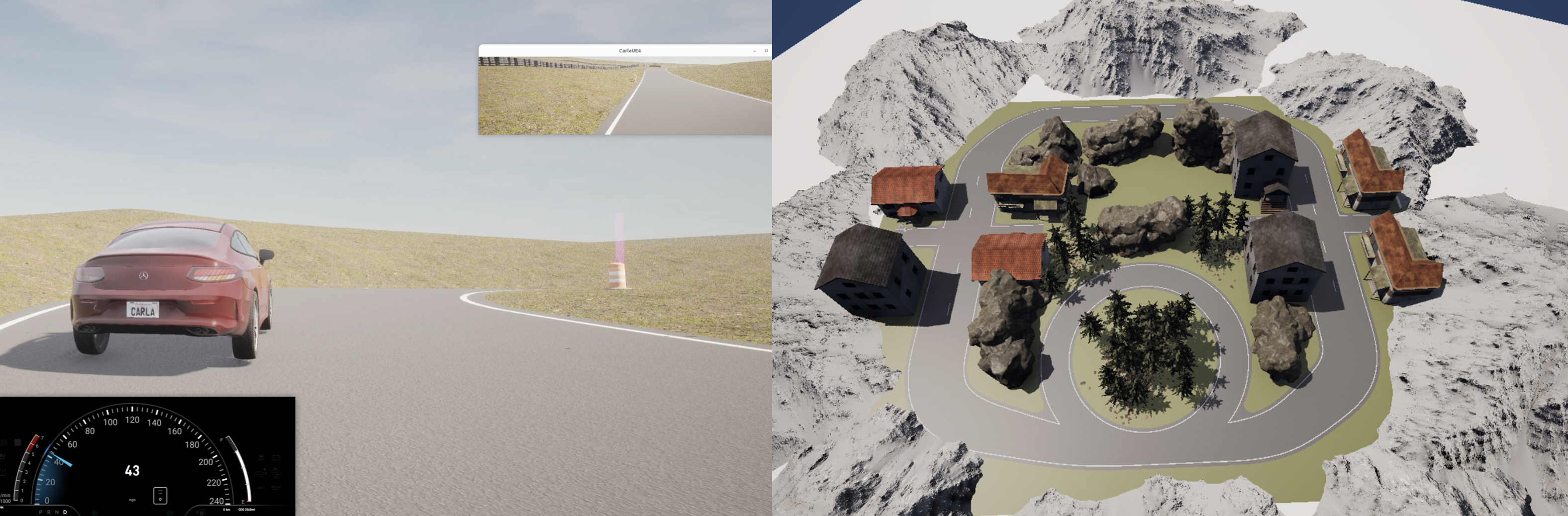}
    \caption{Racing (left) and urban (right) domains.}
    \label{fig:domains}
\end{figure}

We evaluate \dreamtoassist{} across two qualitatively distinct driving domains, illustrated in Fig.~\ref{fig:domains}.  The \textit{racing} domain tests the framework under high-speed, time-pressured conditions with a small intent space. The \textit{urban} domain stress-tests it under surprise-rich, lower-speed conditions with a richer intent space that combines lateral and longitudinal decisions. The urban domain additionally benefits from greater participant familiarity. Our goal is not to compare the two head-to-head, but to draw analytical conclusions for each on its own terms.

\begin{figure}[tb]
    \centering
    \includegraphics[width=0.99\linewidth]{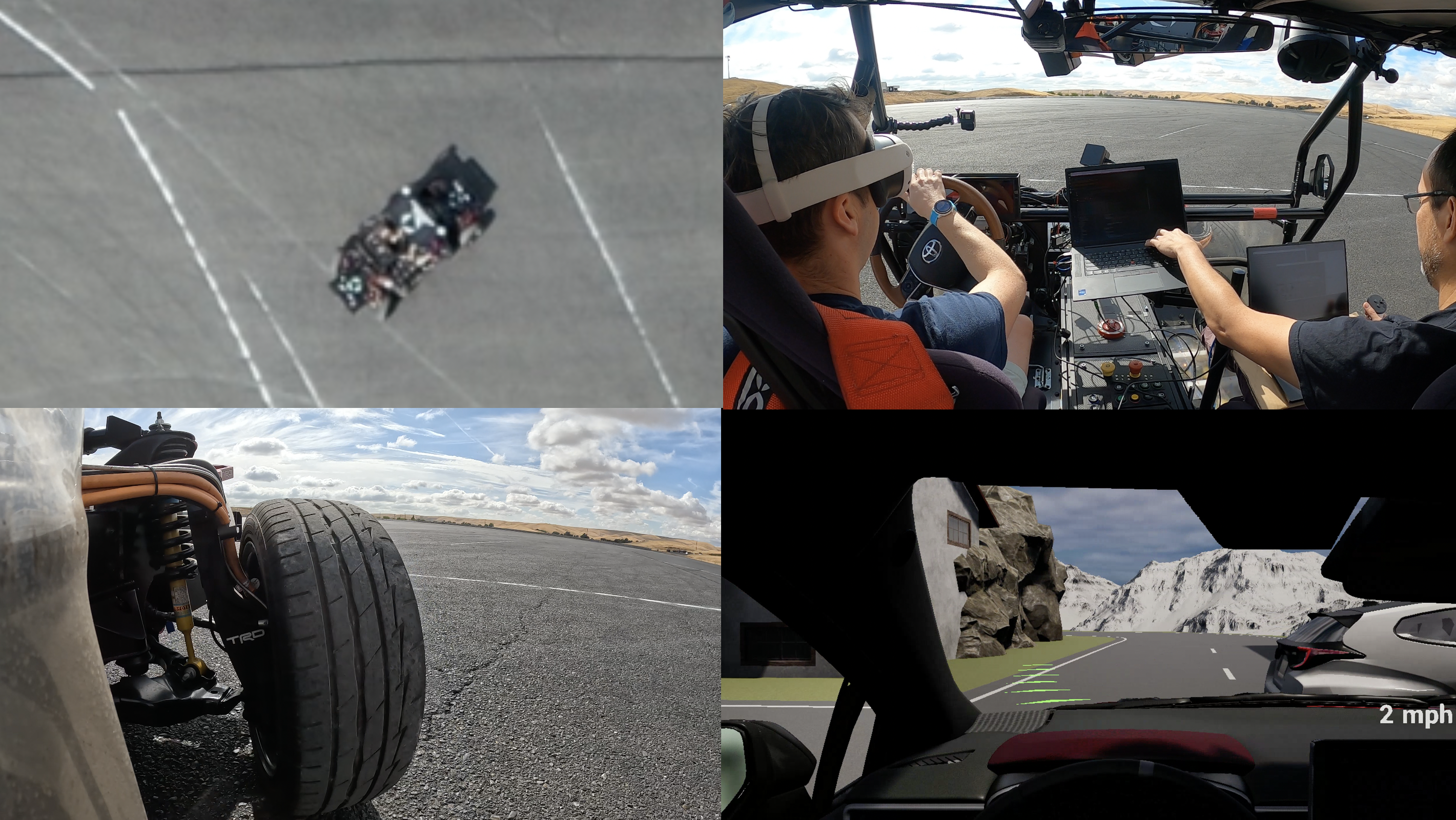}
    \caption{The GRIP physical test platform used for immersive VR-based mixed sim/real testing. %
    }
    \label{fig:grip_platform}
\end{figure}

\begin{figure*}[tb]
    \centering
    \includegraphics[width=0.9\linewidth]{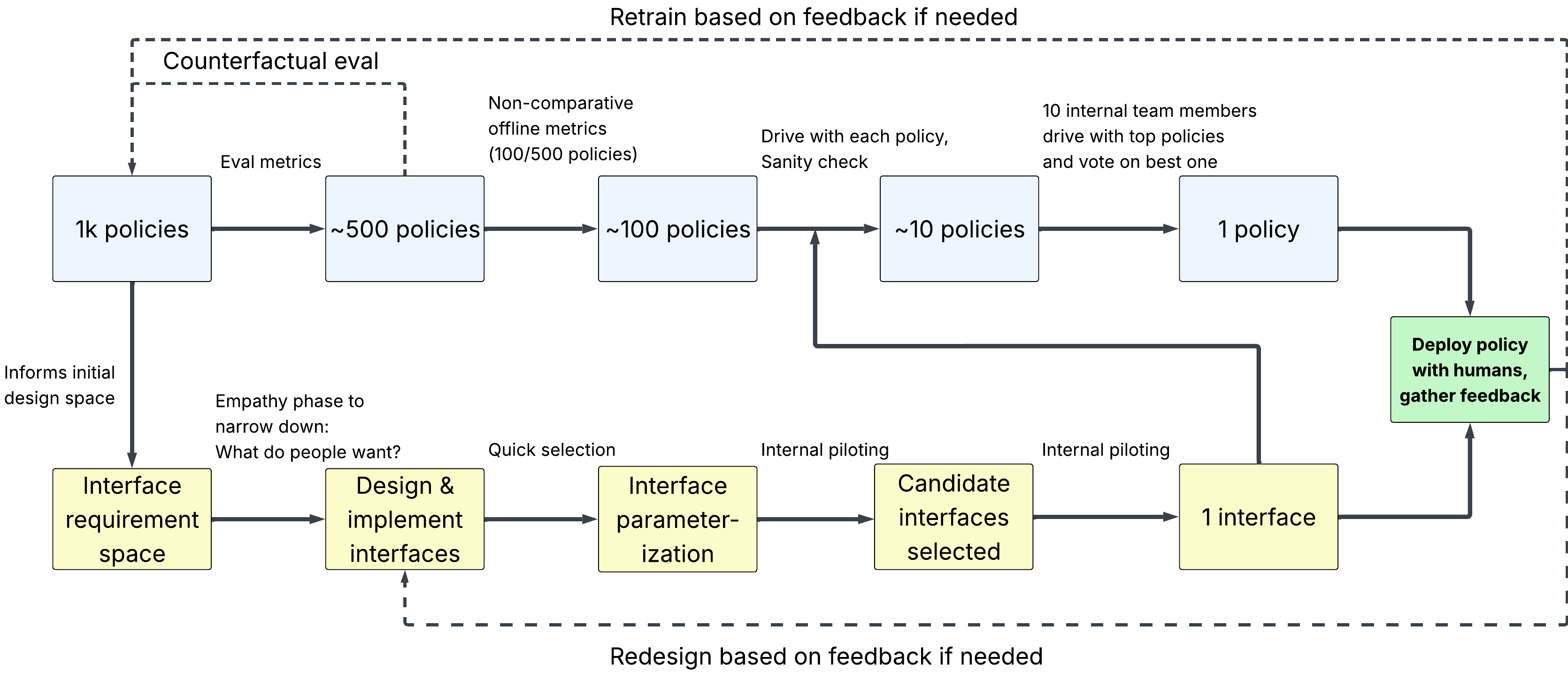}
    \caption{Overall workflow diagram. Notice the combination of iterative refinement and selection of policy choices, as well as the iterative piloting of the overall policy and interface behavior. 
    }
    \label{fig:workflow_diagram}
\end{figure*}

\subsection{Racing Domain}
The racing domain was implemented in the CARLA simulator~\citep{dosovitskiy2017carla} on a portion of the two-mile West track at Thunderhill Raceway in Willows, CA. The track has several characteristic sections that were distilled into different sections where passing could occur: straight and mildly-curved sections with some elevation changes, and highly-curved sections with sharp turns and elevation changes. Both sections offer varied opportunities for one car to overtake another. The intent space comprises two modes: whether to \emph{pass} the lead vehicle or \emph{stay behind} it ($\mathcal{Y} = \{\text{pass}, \text{stay}\}$). The human reward $r_t^y$ encourages proximity to the opponent when passing and distance when staying; the precise formulation is given in Appendix~\ref{sec:policy_interface_design}.

\subsection{Urban Domain}
The urban domain represents a town-like setting with multiple traffic participants, designed to evaluate teaming under everyday hazards. Within the scenario, the ego driver encounters a wrong-way vehicle, a vehicle popping out of a driveway, and a roundabout with merging vehicles. The map is implemented in Unreal Engine and measures 70$\times$70 $m$, dimensioned so the same scenario can also be deployed on physical test skid pads in East Palo Alto, CA and at Thunderhill Raceway.  The intent space is richer than in racing: it combines lateral behaviors (pass on the \textit{left} vs.\ \textit{right}) with longitudinal behaviors (\textit{go} vs.\ \textit{slow down}). The reward structure supports the full product space ($\mathcal{Y} = \{\text{left}, \text{right}\} \times \{\text{go}, \text{slow}, \text{stop}\}$; in the evaluations reported here, only the lateral intents were active ($\mathcal{Y} = \{\text{left}, \text{right}\}$). Domain-specific reward terms are detailed in Appendix~\ref{sec:policy_interface_design}. Compared to racing, the urban domain exercises the policy's intent inference under both spatial and temporal ambiguity.

\subsection{Simulator Infrastructure}
\label{sec:sim_infrastructure}
Both domains run on a common stationary simulator setup. Each rig is outfitted with a Puget Systems workstation with an Intel Core i9 CPU, 128 GB of RAM, and two NVIDIA RTX A6000 Ada GPUs and 38-in monitor.  For the studies, the racing domain used a Fanatec wheel; the urban domain and GRIP used a Logitech G29 wheel. The racing environment is built in CARLA; and the urban environment is built in Unreal Engine, which additionally supports a virtual reality (VR) headset. Policy inference and trajectory/image logging are managed through the Robot Operating System (ROS) in all cases.

A key design choice concerns how control authority is shared between the driver and the policy at the hardware level. Steering is \emph{coupled}: the road wheels always follow the steering wheel, and \dreamtoassist{} communicates its assistive input by applying torque to the steering column.  The driver feels this torque and can comply with or override it at any time.  Throttle and brakes are \emph{uncoupled}: the policy can directly modulate pedal inputs without a corresponding physical force on the pedals and therefore the driver does not physically feel the longitudinal assistance. The interface design choices in Sec.~\ref{sec:deploy} and Appendix~\ref{sec:policy_interface_design} reflect this constraint.

\subsection{GRIP Physical Platform}
\label{sec:grip_platform}
To assess transfer beyond the static simulator, we additionally deploy the urban domain on GRIP~\citep{TRI2024GRIP} (Fig.~\ref{fig:grip_platform}), a higher-fidelity physical test platform. Because GRIP's emulated vehicle dynamics are identical to those used in simulation, only minimal parameter adjustments are required to compensate for differences in steering-wheel hardware. No retraining is performed when going from the simulator to GRIP.

\begin{table*}[htbp]
\centering
\caption{\label{tab:study_overview}%
  Overview of human-subjects studies.
  Each study used a specific system and interface configuration
  detailed in Table~\ref{tab:config_reference}.
}
\footnotesize
\renewcommand{\arraystretch}{1.35}
\resizebox{\linewidth}{!}{%
\begin{tabular}{
  >{\raggedright\arraybackslash}p{3.8cm}
  c
  >{\raggedright\arraybackslash}p{2.8cm}
  >{\raggedright\arraybackslash}p{3.5cm}
  >{\raggedright\arraybackslash}p{5.5cm}
}
\toprule
\textbf{Study} &
$\boldsymbol{N}$ &
\textbf{Environment (map)} &
\textbf{Comparison} &
\textbf{Purpose} \\
\midrule
 
\textbf{Eval~1}: Racing characterization &
24 &
CARLA\newline (Thunderhill) &
\dreamtoassist{}-assisted\newline vs.\ unassisted driving &
Characterize shared control interaction across track geometries; identify failure modes for targeted refinement \\[4pt]
 
\textbf{Eval~2}: Racing refinement &
6 &
CARLA\newline (Thunderhill) &
Refined vs.\ earlier\newline configuration (within-subjects) &
Validate that targeted refinements (data augmentation, output smoothing, reward restructuring) improved the interaction \\[4pt]
 
\textbf{Eval~3}: Urban validation &
10 &
Unreal Engine \newline (CrazyTown) &
Intent-conditioned vs.\ no-intent (double-blind A/B) &
Controlled test isolating the contribution of intent conditioning with the complete \dreamtoassist{} framework \\[4pt]
 
Physical platform (GRIP) &
3 &
GRIP + VR Unreal \newline (CrazyTown) &
Intent-conditioned vs.\ no-intent &
Feasibility demonstration of policy transfer to higher-fidelity motion platform; no retraining \\
 
\bottomrule
\end{tabular}%
}
\end{table*}

\begin{table*}[t]
\centering
\footnotesize
\renewcommand{\arraystretch}{1.25}
\setlength{\tabcolsep}{3pt}
 
\caption{\label{tab:config_reference}%
  Configuration reference for each study (see Table~\ref{tab:study_overview}
  for study overview).
  Sections~III and IV describe all \dreamtoassist{} components in full
  generality; this table records which were active and how they were
  configured in each deployment.
  \cmark\ = active; \xmark\ = not used.
  In \textbf{Eval~1}, \withintentorig{} refers to the configuration shown in the first row. \initialpolicy{} %
  and \refinedpolicy{} are used exclusively in \textbf{Eval~2}.  \withintent{}, \nointent{}, and \nopolicy{} are used exclusively in \textbf{Eval~3} and GRIP.
}
 
\vspace{4pt}

\resizebox{\linewidth}{!}{%
\begin{tabular}{
  >{\raggedright\arraybackslash}p{2.1cm}
  c c c c
  >{\raggedright\arraybackslash}p{2.6cm}
  >{\raggedright\arraybackslash}p{1.0cm}
  >{\raggedright\arraybackslash}p{2.6cm}
  >{\raggedright\arraybackslash}p{2.6cm}
  c
}
\toprule
 
& \multicolumn{6}{c}{\textbf{Policy Training Configuration}}
& \multicolumn{3}{c}{\textbf{Runtime / Interface Configuration}} \\
 
\cmidrule(lr){2-7} \cmidrule(lr){8-10}

\textbf{Study} &
\rotatebox{70}{\makecell[l]{Human data\\augmentation}} &
\rotatebox{70}{\makecell[l]{FCP output\\smoothing}} &
\rotatebox{70}{\makecell[l]{Intent class\\rebalancing}} &
\rotatebox{70}{\makecell[l]{Longitudinal}} &
\makecell[c]{\textbf{Reward + Intents}} &
\rotatebox{70}{\makecell[l]{Safety\\layer}} &
\makecell[c]{\textbf{Trajectory}\\\textbf{visualization}} &
\makecell[c]{\textbf{Haptics}\\\textbf{mode}} &
\rotatebox{70}{\makecell[l]{Output\\smoothing}} \\
 
\midrule
 
\textbf{Eval~1:} Racing characterization ($N{=}24$) &
\xmark &
\xmark &
\xmark &
\xmark &
Baseline task rewards (Appendix~\ref{sec:policy_interface_design}); default pass/progress scalars.
$\mathcal{Y} = \{\text{pass}, \text{stay}\}$. &
\xmark &
Line by proximity to opponent; color by accel/decel &
Steering setpoint of human+policy; direct pedal blending &
\xmark \\

\addlinespace[6pt]
 
\textbf{Eval~2:} Racing refinement ($N{=}6$) &
\cmark\ (20\%) &
\cmark &
\xmark &
\xmark &
Augmented stay-behind and pass rewards (\eqref{eq:rew_pass_stay}--\eqref{eq:rew_total_pass_prog}). 
$\mathcal{Y} = \{\text{pass}, \text{stay}\}$. &
\xmark &
Line by proximity to opponent; color by accel/decel &
Steering setpoint of human+policy; direct pedal blending &
\xmark \\

\addlinespace[6pt]
 
\textbf{Eval~3:} Urban validation ($N{=}10$) &
\xmark &
\cmark &
\cmark &
\xmark &
Refined collision, boundary, speed, proximity features, left/right passing rewards (Appendix~\ref{sec:policy_interface_design}).
$\mathcal{Y} = \{\text{left}, \text{right}\}$. &
\cmark\ (traj.~viz. only) &
Line by policy actions (alignment); color by accel/decel; windowed smoothing &
Steering setpoint of human+policy; direct pedal blending &
\cmark \\

\addlinespace[6pt]
 
GRIP platform ($N{=}3$) &
\xmark &
\cmark &
\cmark &
\cmark &
As urban + longitudinal behaviors (stop/go rewards in Appendix~\ref{sec:policy_interface_design}).
$\mathcal{Y} = \{\text{left}, \text{right}\} \times
\{\text{go}, \text{slow}, \text{stop}\}$. &
\cmark\ (traj.~viz. only) &
As urban validation &
As urban; minimal parameter adjustment for platform hardware &
\cmark \\
 
\bottomrule
\end{tabular}%
}
\end{table*}

\section{Policy \& Interface Design Process}
\label{sec:deploy}
Deploying a human-agnostic policy with real users requires going from a population of trained candidates to a single deployed system: both a policy and an interface that pairs with it. We approach this via two parallel iterative pipelines, summarized in Fig.~\ref{fig:workflow_diagram}, consisting of a \textit{policy down-selection} pipeline that narrows a large pool of trained candidates to one, and an \textit{interface design} pipeline that shapes how the chosen policy communicates with and acts on the driver. The term \textit{interface} refers to any mechanism or modality used for communicating the policy's decision to the human, for example, trajectory visualizations of future rollouts, steering wheel torque modulation to convey the strength of assistance etc. \textit{Interface design} can involve multiple decisions, for example, gain parameters for the torque feedback, choice of trajectory visualization \textit{etc}. Both pipelines share a common backbone of iterative refinement informed by piloting, and we treat the joint output as a contribution of this work.

The two pipelines are coupled. Interface design can be initiated against any candidate policy, but is refined as the policy field narrows; conversely, late-stage policy voting uses a near-final interface so that participants are reacting to the policy itself, not the display. We applied this full workflow in the urban domain, where the variety of everyday driving situations argues for diverse opinions and we used voting at the late stages of policy selection. In the more objective-driven racing domain, we relied on offline metrics alone for the policy side and did not use voting.

\subsection{Selection Spaces}
Up to this point, candidate policies have only been evaluated against simulated humans under simulated conditions. To bridge to deployment, we apply a three-stage funnel that progressively trades evaluation cost for evaluation fidelity. Hyperparameter sweeps and random seeds typically yield on the order of 1K candidates; the funnel narrows this to a single deployed policy. Stage 1 reduces the number of policies to ${\sim}500$; Stage 2 to ${\sim}100$; Stage 3 human evaluation selects from 5--12 final policies.

\paragraph*{Stage 1: Non-comparative offline metrics.} Each candidate is scored on training-time statistics from fictitious co-play: task progress, average speed, collision rate, intent-classification accuracy, and intervention sparsity. Candidates falling below pre-set thresholds are eliminated. This stage is essentially free per candidate and absorbs the bulk of the field.

\paragraph*{Stage 2: Counterfactual log-replay evaluation.} Surviving policies are scored against log replays of real human drivers. For each logged human input sequence, we compute what would have happened with the candidate policy active (assisted) and compare to the original unassisted trajectory.  Two metrics target deployment risks that are not visible in training statistics: \textit{collision propensity}, the dot-product of the assistive steering and throttle vectors with the relative direction of ego and opponent vehicles when within a proximity threshold; and \textit{track-orthogonality}, the perpendicularity of the assistive control vector with respect to the track direction, which captures the policy's propensity to spin the driver out or off-track. Because these run against fixed logs, two or more policies can be compared on identical human inputs.

\paragraph*{Stage 3: Human-in-the-loop evaluation.} Stages 1–2 may still leave ${\sim}100$ candidates which are too numerous for a human study. We therefore run a two-round internal review in the simulator: in the first round, a single rater checks each surviving candidate against pre-defined pass/fail criteria %
in the second round, a different rater re-rates the remainder and selects forward. For the urban domain, we additionally run a vote among ten internal team members on the top survivors, using a near-final interface so voters respond to policy behavior rather than display differences.

\subsection{Interface Design Process}
\label{sec:interface_design_process}
In parallel with policy selection, we ran an iterative human-machine-interface design cycle~\citep{wickens2004introduction,mehrotra2022human}, which is standard in the HMI literature. Because learned policies are not guaranteed to be safe~\citep{oh2025safety}, the safety filter was applied to the trajectory visualization to ensure that displayed paths remain within safe bounds (Sec.~\ref{sec:safety_layer}).  The workflow is shown in the bottom row of Fig.~\ref{fig:workflow_diagram}.  We begin with an empathy phase that scopes the requirement space to define the kinds of cues, modalities, and timing drivers want from a shared controller. Candidate interface concepts are then implemented and parameterized (line color, line length, audio cue taxonomy, haptic gain) and run through repeated piloting sessions with both internal and external participants. Late in the cycle, surviving candidates are narrowed via a voting procedure with pilots, again paired with a near-final policy. 

Across piloting and formal evaluation sessions, participants offered open feedback on the relative salience of the two primary feedback channels. Visual trajectory cues were generally preferred for spatial guidance:
\begin{quote}
``I like the green line. It felt like a built-in GPS of some sort. I didn't want to go in that exact direction, like when the car was pulling out, but liked how the steering wheel was nudging[\ldots] it was mapping out what I was doing, plus what's good for me.'' (P05)
\end{quote}
\noindent While haptic feedback was valued for its immediacy, some participants found it intrusive at higher intensities:
\begin{quote}
``I liked the visuals better than the haptics, because the haptics were more startling and distracting.'' (P06)
\end{quote}
\noindent These observations are consistent with findings in the driving HMI literature, where visual cues are preferred for planning and spatial awareness while haptic cues are preferred for urgent, directional guidance~\citep{ploch2017comparing, Balachandran2016-sl}. We did not conduct a formal modality comparison study; the feedback above reflects informal observations that informed our interface tuning.

Table~\ref{tab:HF_factors} lists the full set of factors considered, including those rejected during piloting (e.g., audio utterance cues), and Appendix~\ref{sec:policy_interface_design} describes the surviving design choices in detail. The output of this cycle, together with the deployed policy from Sec.~\ref{sec:deploy} defines the joint system used in the human evaluations of Sec.~\ref{sec:evaluation}.
While we instantiate the proposed pipeline in terms of shared control for driving, the three-stage structure (automated screening, counterfactual replay, human-in-the-loop vetting) applied more broadly to any domain where learned policies must be vetted before deployment with humans, e.g. manipulation, navigation, and assistive robotics.

%% file: 6_human_evaluation.tex
\FloatBarrier
\begin{table}[!t]
\centering
\footnotesize
\renewcommand{\arraystretch}{1.2}
\setlength{\tabcolsep}{3pt}
\caption{\label{tab:eval1_subjective}%
  Post-drive survey ratings for \textbf{Eval~1} ($N = 24$),
  Likert 1--7 (Neutral = 4).
  One-sample $t$-tests against neutral (two-tailed, $df = 23$).
  $\uparrow$~higher is better;
  $\downarrow$~reverse-coded.
  ***\,$p < .001$; **\,$p < .01$.
  Straight vs.\ turn breakdowns: intent recognition higher on
  straights ($p = .001$); helpfulness above neutral on straights
  ($p = .054$) but below on turns ($p = .025$); safety above neutral
  on straights ($p < .001$) but not turns ($p = .057$).%
}
\resizebox{\columnwidth}{!}{%
\begin{tabular}{
  >{\raggedright\arraybackslash}p{2.0cm}
  c c c
  r @{\hspace{4pt}} 
  c
}
\toprule
 &
\textbf{Str.} &
\textbf{Turn} &
\textbf{All} &
\multicolumn{2}{c}{\textbf{vs.\ Neutral}} \\
\cmidrule(lr){2-2} \cmidrule(lr){3-3} \cmidrule(lr){4-4} \cmidrule(lr){5-6}
\textbf{Question} &
$M \pm SE$ &
$M \pm SE$ &
$M \pm SE$ &
$t(23)$ & \multicolumn{1}{c}{$p$} \\
\midrule
 
AI knew my intent to pass\,$\uparrow$ &
$5.49 \pm .09$ &
$5.09 \pm .13$ &
$5.33 \pm .08$ &
$10.63$ & ${<}\,.001$*** \\
 
AI helped me perform the maneuver\,$\uparrow$ &
$4.01 \pm .13$ &
$3.16 \pm .14$ &
$3.67 \pm .10$ &
$0.44$ & $.666$ \\
 
AI's action was safe\,$\uparrow$ &
$4.55 \pm .11$ &
$3.80 \pm .14$ &
$4.25 \pm .09$ &
$3.60$ & $.002$** \\
 
Better off without AI\,$\downarrow$ &
$4.82 \pm .11$ &
$5.12 \pm .12$ &
$4.94 \pm .08$ &
$3.57$ & $.002$** \\
 
\bottomrule
\end{tabular}%
}
\end{table}

\section{Human Evaluations}
\label{sec:evaluation}
We conducted three human-subjects studies and one preliminary on-vehicle deployment, as summarized in Table~\ref{tab:study_overview}, with study-specific configurations in Table~\ref{tab:config_reference}. The three studies build upon one another. \textbf{Eval 1} (Racing Characterization) characterizes how ~\dreamtoassist{} performs against an unassisted baseline in the racing domain. \textbf{Eval 2} (Racing Policy Refinement) validates targeted refinements (human-data augmentation, action smoothing, reward restructuring, safety filtering), on the same domain. It provides directional evidence that the refined configuration improved subjective ratings and turn-segment performance relative to the original configuration, though the sample (N=6) is underpowered for definitive claims. \textbf{Eval 3} (Urban Controlled Validation) applies the refined framework to the urban domain, and runs a double-blind A/B comparison to isolate the contribution of intent conditioning. Lastly, we report a preliminary on-vehicle deployment of the urban scenario on the GRIP physical platform.

\subsection{Common Methodology}
The following procedures were applied across all three studies. Participants filled out an informed-consent form covering study risks (eye strain, motion sickness) before entering the simulator room.
Internal participants were compensated with \$80 USD worth of company prizes (such as sweatshirts and tote bags); external participants received \$150 USD. Each study (except GRIP) ran on the simulator hardware, software stack, and ROS-based logging described in Sec.~\ref{sec:sim_infrastructure}.  Unless otherwise noted, subjective ratings were collected on 7-point Likert scales. In addition, participants provided an overall policy rating on a 1–10 scale reflecting their global impression of the AI's assistance quality. Study-specific demographics, inclusion criteria, and procedural details are reported within each subsection.  Additional details for \textbf{Eval 1} and \textbf{Eval 3} are reported in Appendix~\ref{sec:detailed_methods}.
This study was conducted under the oversight of WCG IRB on behalf of Toyota Research Institute (IRB Tracking Number: 20232718, Study Number: 1356002). Informed consent was obtained from all participants prior to their participation.

\subsection{Evaluation 1: Racing Characterization ($N=24$)}
\label{sec:eval1}
Our goal is to characterize how \dreamtoassist{} interacts with human drivers across two qualitatively different track segments (straightaway and turn), and identify failure modes for targeted refinement. We hypothesized that the policy would improve overtake success and opponent clearance relative to an unassisted baseline.
We recruited 33 participants ($N=24$ after exclusions; see Appendix~\ref{sec:detailed_methods} for full demographics).

\paragraph*{Study Design}
Within-subjects comparison of \dreamtoassist{}-assisted vs. unassisted driving in CARLA on the 2-mile West track at Thunderhill Raceway (see row 1 of Table~\ref{tab:study_overview}).  Familiarization consisted of one observation lap, five laps coached by the rule-based AI coach of~\cite{kaufman2024effects}, and five laps without coaching. The main task was structured into eight overtake scenarios of increasing difficulty using a staircase approach~\citep{cornsweet1962staircase}.  Each block contained two follow laps (to observe opponent behavior) and two overtake-attempt laps between marked purple gates. Participants advanced to the next scenario only upon successful
completion of the task (see Appendix~\ref{sec:detailed_methods}).  Each scenario was completed both without and with the policy.  
A video montage of representative trials is included in the
supplementary material (Video~1).

\paragraph*{Measures}
\textbf{Objective:} Overtake rate, mean overtake speed, minimum opponent clearance, off-track rate, spinout rate, collision rate, lane-boundary distance, and torque-alignment events.
\textbf{Subjective:} Per-policy 7-point Likert ratings on whether: (1) the AI knew my intent to pass, (2) the AI helped perform the maneuver, (3) the AI's action was safe, and (4) I would have been better off without the AI. In addition, an overall 1–10 rating and open-ended qualitative feedback was collected after the driving session.

\paragraph*{Objective Findings}
We tested six secondary safety metrics (spinout rate, off-track rate, collision rate, right-side pass rate, mean overtake speed, and minimum opponent clearance (Tables~\ref{tab:outcomes_by_track}--~\ref{tab:overtake_quality_track}).  On the straight segments, \dreamtoassist{} increased per-participant overtake success from 80\% to 89\% ($p = .069$, $n = 23$ participants with trials in both conditions), increased overtake speed, and increased minimum opponent clearance.  
On turns, \dreamtoassist{} reduced per-participant overtake success from 51\% to 21\% ($p < .001$, $n = 23$) and reduced minimum opponent clearance from 1.99\,m to 1.01\,m (also significant)..  Other safety metrics (spinouts, off-track, collisions) showed no meaningful differences between conditions (Table~\ref{tab:overall_outcomes}). We identified, via inspection of the assistive control inputs, that over-steering by the policy on turns was the likely cause of the turn regression.

\begin{table}[htbp]
\caption{Behavioral Outcomes by Track Type: \withintentorig{} vs.\ \nopolicy{} ($N = 24$). Values are proportions pooled over trials; the paired tests in the text use per-participant proportions.}
\label{tab:outcomes_by_track}
\centering
\resizebox{\columnwidth}{!}{
\begin{tabular}{p{2.6cm}lccc}
\toprule
Outcome & Track & \nopolicy{} & \withintentorig{} & $\Delta$ \\
        &       & $M \pm \mathit{SE}$ & $M \pm \mathit{SE}$ & \\
\midrule
\multirow{2}{=}{Overtake Rate $\uparrow$}
  & Straight & $0.810 \pm .034$ & $\mathbf{0.899 \pm .027}$ & $+8.9\%$  \\
  & Turn     & $\mathbf{0.547 \pm .058}$ & $0.274 \pm .046$ & $-27.3\%$ \\[2pt]
\multirow{2}{=}{Spinout Rate $\downarrow$}
  & Straight & $0.146 \pm .030$ & $\mathbf{0.132 \pm .030}$ & $-1.4\%$  \\
  & Turn     & $\mathbf{0.293 \pm .053}$ & $0.326 \pm .048$ & $+3.3\%$  \\[2pt]
\multirow{2}{=}{Off-Track Rate $\downarrow$}
  & Straight & $0.073 \pm .022$ & $\mathbf{0.070 \pm .023}$ & $-0.3\%$  \\
  & Turn     & $\mathbf{0.120 \pm .038}$ & $0.137 \pm .035$ & $+1.7\%$  \\[2pt]
\multirow{2}{=}{Collision Rate $\downarrow$}
  & Straight & $\mathbf{0.015 \pm .010}$ & $0.023 \pm .013$ & $+0.9\%$  \\
  & Turn     & $\mathbf{0.013 \pm .013}$ & $0.032 \pm .018$ & $+1.8\%$  \\[2pt]
\multirow{2}{=}{Right-Side Pass}
  & Straight & $0.223 \pm .038$ & $0.344 \pm .043$ & $+12.1\%$ \\
  & Turn     & $0.735 \pm .064$ & $0.756 \pm .065$ & $+2.1\%$  \\
\bottomrule
\end{tabular}%
}
\end{table}

\paragraph*{Subjective Findings}
Likert ratings (Fig.~\ref{tab:eval1_subjective}, Table~\ref{tab:eval1_subjective}, were analyzed through one-sample and paired t-tests, all of which were two-tailed, $N=24$) showed that participants rated the policy as understanding their intent significantly above neutral overall ($M = 5.29$, $t(23) = 10.63$, $p < .001$), with ratings significantly higher on straights than turns ($t(23) = 3.68$, $p = .001$).  Perceived helpfulness did not differ from neutral overall ($M = 3.59$, $t(23) = 0.44$, $p = 0.666$); straights trended above neutral ($M = 4.00$, $t(23) = 2.04$, $p = .054$), while turns fell significantly below ($M = 2.99$, $t(23) = -2.39$, $p = .025$).  Safety was rated significantly above neutral overall ($M = 4.10$, $t(23) = 3.60$, $p = .002$) and on straights ($M = 4.41$, $t(23) = 4.38$, $p < .001$), though lower on turns ($M = 3.64$, $t(23) = 3.57$, $p = .002$).

\paragraph*{Implications}
The policy is rated as understanding intent and as helpful and safe on straightaways, but degrades on turns due to a dynamics-complexity boundary.  This motivates the targeted refinements evaluated in the next section.

\begin{table}[htbp]
\caption{Overall Behavioral Outcomes: \withintentorig{} vs.\ \nopolicy{} ($N = 24$). Values are proportions pooled over trials; the paired tests in the text use per-participant proportions.  Note that overall this initial policy \textit{inhibited} the driver.}
\label{tab:overall_outcomes}
\centering
\resizebox{\columnwidth}{!}{%
\begin{tabular}{lccc}
\toprule
Outcome & \nopolicy{} & \withintentorig{} & $\Delta$ \\
        & $M \pm \mathit{SE}$ & $M \pm \mathit{SE}$ & \\
\midrule
Overtake Rate $\uparrow$   & $\mathbf{0.721 \pm .031}$ & $0.630 \pm .032$ & $-9.0\%$  \\
Spinout Rate $\downarrow$    & $\mathbf{0.195 \pm .027}$ & $0.209 \pm .027$ & $+1.3\%$  \\
Off-Track Rate $\downarrow$  & $\mathbf{0.088 \pm .019}$ & $0.104 \pm .020$ & $+1.6\%$  \\
Collision Rate $\downarrow$  & $\mathbf{0.014 \pm .008}$ & $0.026 \pm .011$ & $+1.2\%$  \\
Right-Side Pass & $0.376 \pm .037$ & $0.465 \pm .038$ & $+8.9\%$  \\
\bottomrule
\end{tabular}%
}
\end{table}

\begin{table}[htbp]
\caption{Overtake Quality Metrics: \withintentorig{} vs.\ \nopolicy{} (Overtaking Trials Only). Values are means $\pm$ SE over overtaking trials.}
\label{tab:overtake_quality}
\centering
\setlength{\tabcolsep}{4pt}
\resizebox{\columnwidth}{!}{%
\begin{tabular}{lccc}
\toprule
Metric & \nopolicy{} & \withintentorig{} & $\Delta$ \\
       & $M \pm \mathit{SE}$ & $M \pm \mathit{SE}$ & \\
\midrule
Mean Overtake Speed (m/s) $\uparrow$ & $\mathbf{26.39 \pm .29}$ & $26.27 \pm .28$ & $-0.13$ \\
Min Opp. Clearance (m) $\uparrow$  & $1.62  \pm .10$ & $\mathbf{1.77  \pm .48}$ & $+0.14$ \\
\bottomrule
\end{tabular}%
}
\end{table}

\begin{table}[t]
\caption{Overtake Quality Metrics by Track Type: \withintentorig{} vs.\ \nopolicy{} (Overtaking Trials Only)}
\label{tab:overtake_quality_track}
\centering
\setlength{\tabcolsep}{3.5pt}
\resizebox{\columnwidth}{!}{
\begin{tabular}{p{3.9cm}lccc}
\toprule
Metric & Track & \nopolicy{} & \withintentorig{} & $\Delta$ \\
       &       & $M \pm \mathit{SE}$ & $M \pm \mathit{SE}$ & \\
\midrule
\multirow{2}{=}{Mean O/T Speed (m/s) $\uparrow$}
  & Straight & $\mathbf{27.89 \pm .22}$ & $27.44 \pm .22$ & $-0.46$ \\
  & Turn     & $\mathbf{22.67 \pm .51}$ & $21.06 \pm .36$ & $-1.62$ \\[2pt]
\multirow{2}{=}{Min Opp. Clearance (m) $\uparrow$}
  & Straight & $1.50 \pm .11$ & $\mathbf{1.97 \pm .59}$ & $+0.47$ \\
  & Turn     & $\mathbf{1.99 \pm .21}$ & $1.01 \pm .22$ & $-0.98$ \\
\bottomrule
\end{tabular}%
}
\end{table}

\subsection{Evaluation 2: Racing Policy Refinement ($N = 6$)}
\label{sec:eval2}
The goal for this evaluation is to validate that the refinements introduced after \textbf{Eval 1}, i.e. human-data augmentation, output smoothing, and reward restructuring, improve the interaction over the original \dreamtoassist{} configuration, with focus on turns.  As a within-subjects pilot, this study targets directional trends rather than statistical significance.  The reward changes targeted the pass/stay formulation specifically: asymmetric scaling of pass vs.\ stay rewards and a bounded progress term to prevent mode collapse toward always-passing behavior (\eqref{eq:rew_pass_stay}--\eqref{eq:rew_total_pass_prog}, Appendix~\ref{sec:policy_interface_design}).  The intent space remained $\mathcal{Y} = \{\text{pass}, \text{stay}\}$. 
For this study, we recruited $N=6$ internal participants.

\paragraph*{Study Design}
Within-subjects comparison of \refinedpolicy{} against \initialpolicy{} on the racing domain (see row 2 of Table~\ref{tab:study_overview}). The procedure was a shorter version of \textbf{Eval~1}, retaining the staircase scenarios. Each participant drove every scenario with each policy.

\paragraph*{Measures}
Objective measures matched \textbf{Eval~1}. Subjective measures included per-policy Likert ratings of likability and helpfulness, plus per-trial preference (which policy was preferred for each scenario) and open-ended likes/dislikes.

\begin{figure}
    \centering
    \includegraphics[width=0.7\linewidth]{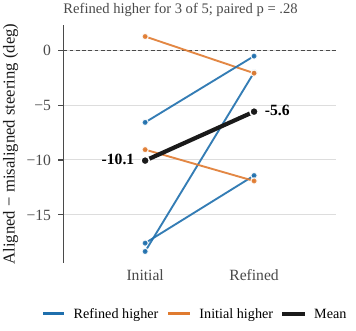}
    \caption{Aligned minus misaligned AI steering per participant under \initialpolicy{} and \refinedpolicy{} ($n = 5$; P106 excluded for missing steering data). Higher values mean more steering in the driver's intended direction. Thin lines are participants, the thick line the mean; paired $t$-test $p = .28$.}
    \label{fig:eval_2_torque_alignment}
\end{figure}

\paragraph*{Objective Findings}
We examine whether the four refinements reduced the excessive steering torque on turns that we identify as the primary failure mode in \textbf{Eval~1}. Figure~\ref{fig:eval_2_torque_alignment} shows the torque-magnitude difference $\Delta\tau$ \eqref{eq:torque_difference} for each condition. Under both configurations, the policy applied more intent-opposing than intent-aligned steering on average, consistent with the over-steering failure mode on turns. Under \refinedpolicy{}, intent-opposing steering decreased slightly (31.2 to 29.6\,deg) and the aligned-minus-opposing difference rose from $-10.1$ to $-5.6$\,deg, higher for 3 of 5 participants (P106 excluded: missing \refinedpolicy{} steering data).  Consistent with the pilot’s aim of directional trends, neither change is significant at this sample size (paired $t$-tests, $p = .68$ and $p = .28$, $n = 5$), which we take as directional evidence that action smoothing and reward restructuring reduced the policy's tendency to fight the driver.

\begin{figure}
    \centering
    \includegraphics[width=0.99\linewidth]{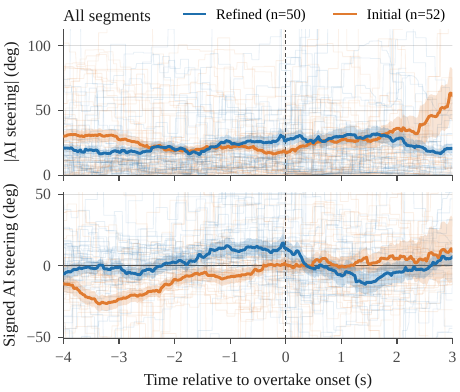}
    \caption{AI steering around overtake onset ($t = 0$), all participants and segments (\initialpolicy{}: 52 passing events; \refinedpolicy{}: 50). Top: steering magnitude. Bottom: signed steering. Lines show mean $\pm$ SEM; individual events dimmed.}
    \label{fig:eval_2_torque_traces}
\end{figure}

Averaged AI steering traces time-aligned to overtake onset (Fig.~\ref{fig:eval_2_torque_traces}) differ most on turns: in the 4--3\,s before the overtake, \initialpolicy{} applies roughly twice the steering magnitude of \refinedpolicy{} (about 50 vs.\ 25\,deg), consistent with the over-steering that characterized \initialpolicy{}. The traces show raw signed steering, so they do not by themselves separate intent-aligned from intent-opposing input.

\begin{figure}
    \centering
    \includegraphics[width=0.7\linewidth]{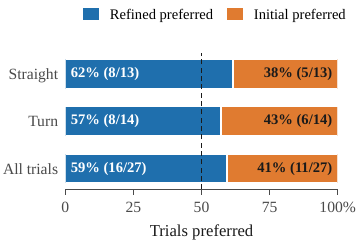}
    \caption{Policy preferences between the initial vs. updated policy faceted by position where the overtake occurred.}
    \label{fig:policy_refinement_straights_v_turns_pref}
\end{figure}

\paragraph*{Subjective Findings}
Likability was numerically higher for \refinedpolicy{} ($M = 3.31$, $\mathit{SE} = 0.39$)  than for \initialpolicy{} ($M = 2.95$, $\mathit{SE} = 0.42$).  Helpfulness ratings were comparable across versions (refined: $M = 2.85$, $\mathit{SE} = 0.20$; $t(5) = -0.15$, $p = .887$, $d = -0.06$; initial: $M = 2.80$, $\mathit{SE} = 0.33$).  At the trial level ($N = 27$ trials, using a two-sided binomial test), participants preferred \refinedpolicy{} on 59.3\% of trials (16 of 27), compared to 40.7\% for the characterization policy (11 of 27), with $p = 0.442$.  Preference did not favor turns: \refinedpolicy{} was preferred on 8 of 14 turn trials (57\%) and 8 of 13 straight trials (62\%) (Fig.~\ref{fig:policy_refinement_straights_v_turns_pref}). Likability was numerically higher for \refinedpolicy{} (Table~\ref{tab:eval2_subjective}), consistent with the refinements targeting the turn-specific failure mode identified in \textbf{Eval 1}.

\paragraph*{Implications}
The pilot supports a directional improvement attributable to the refinements, particularly on turns. Comparing turns in \textbf{Eval~1}, the original policy had learned human dynamics associated with fictitious human partners, which tended to steer more aggressively with higher repeatability in turns than true humans. We hypothesize that the simulated humans had higher repeatability on turns than real humans, so the world model learned overly aggressive turn dynamics. This caused excessive steering over-compensation just before the turn, which degraded safety and overtake success. Human data augmentation exposed the world model to more in-distribution human behaviors, reducing the domain gap in terms of dynamics of the human partner. The additional reward restructuring penalized mode collapse to a single intent, thereby allowing the policy to distinguish between intents especially in turns where intent is ambiguous. Together, these allowed the policy to steer less aggressively in dynamically-complex conditions, which may explain the higher likability of \refinedpolicy{}.

With the racing domain validated, \textbf{Eval 3} adopts a qualitatively different domain (urban driving with a richer intent space) to test generalization and the contribution of intent-conditioning specifically.

\begin{figure}
    \centering
    \includegraphics[width=0.99\linewidth]{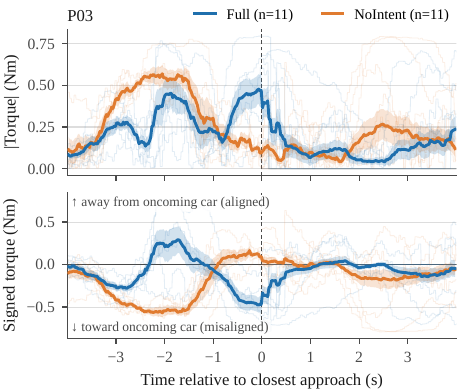}
    \caption{Averaged torque traces for participant 03 in the oncoming car event.  About 2.3 to 1.5\,s before closest approach ($t = 0$), \withintent{} steers away from the oncoming car, in line with the driver's avoidance, while \nointent{} steers toward it (bottom trace).  In the final second the pattern reverses: \withintent{} turns toward the car and \nointent{} is near neutral.}
    \label{fig:eval_3_torque_traces}
\end{figure}

\subsection{Evaluation 3: Urban Controlled Validation ($N = 10$)}
\label{sec:eval3}
The goal of this evaluation is to isolate the contribution of intent conditioning by comparing the full \dreamtoassist{} framework (\withintent{}) against an ablated version without an intent predictor (\nointent{}) in an urban setting. An additional \nopolicy{} condition serves as an unassisted baseline. We hypothesized that intent-conditioning would produce greater torque alignment with each driver's lateral pass choice (to go left or right around an oncoming car) and reduce collisions.  
\withintent{} incorporates all refinements from \refinedpolicy{}{} plus intent-class rebalancing and refined proximity features adapted to the urban domain (Table~\ref{tab:config_reference}, row~3). The intent space for this evaluation was $\mathcal{Y} = \{\text{left}, \text{right}\}$: the policy has to infer whether the driver intended to pass the hazard on the left or right, to shape its assistance accordingly. Left/right reward terms penalize lateral positions inconsistent with the inferred intent (Appendix~\ref{sec:policy_interface_design}).  \nointent{} uses the same task rewards and policy architecture but without the intent predictor, so its assistance is not conditioned on the driver's choice of lateral direction.
For this study, we recruited 12 participants ($N=10$ after exclusions; see Appendix~\ref{sec:detailed_methods} for full demographics).

\paragraph*{Study Design}
We conducted a double-blind within-subjects A/B comparison of \withintent{} vs. \nointent{}, with an additional \nopolicy{} condition (see row 3 of Table~\ref{tab:study_overview}).  The simulator was outfitted with a G29 wheel and the policy communication was via ROS.  Since the study only examined lateral behaviors, we include two opponent-vehicle scenarios: a vehicle backing out of a driveway, and a wrong-way oncoming vehicle.  After familiarization (four town-map laps and practice with the two scenarios), participants completed a three-phase policy onboarding (haptics-only, trajectory-only, combined) with verbal comprehension checks before advancing. Each participant then drove six laps with each of the three conditions in randomized order, alternating left and right passes around the oncoming car each lap. Surveys were administered between blocks.
A video montage illustrating the \withintent{} condition is provided in the supplementary material (Video~2).

\paragraph*{Measures}
\textbf{Objective:} Torque-alignment events (overall and per-participant) Eq.~\eqref{eq:alignment_events}, torque-magnitude difference (good vs. bad alignment) Eq.~\eqref{eq:torque_difference}, and collision rate.
\textbf{Subjective:} Per-policy Likert ratings on directionality, timing, adapting direction, fight direction, safety, teaming, and an overall 1–10 rating.  Objective torque analyses use all 12 sessions (11 with both AI conditions, since no \nointent{} trials were logged for P04); subjective analyses exclude P04 and P12.

\begin{figure*}[htbp]
    \centering
    \par\smallskip
    \begin{subfigure}[b]{0.48\textwidth}
        \centering
        \includegraphics[width=\textwidth]{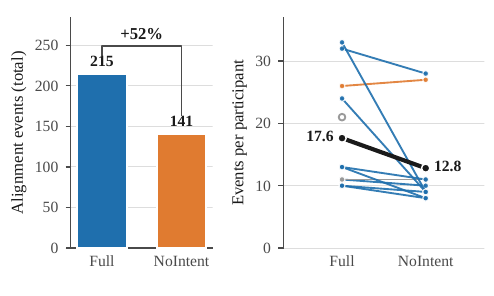}
        \caption{Overall torque alignment counts.}
        \label{fig:torque_alignment}
    \end{subfigure}
    \begin{subfigure}[b]{0.25\textwidth}
        \centering
        \includegraphics[width=\textwidth]{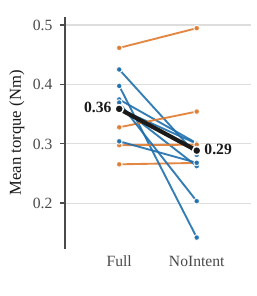}
        \caption{Torque magnitude statistics.}
        \label{fig:torque_magnitude_stats}
    \end{subfigure}
    \begin{subfigure}[b]{0.25\textwidth}
        \centering
        \includegraphics[width=\textwidth]{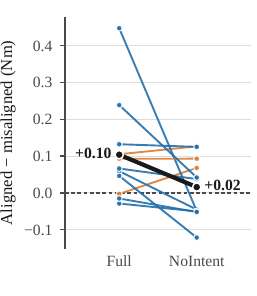}
        \caption{Torque magnitude difference.}
        \label{fig:torque_magnitude_difference}
    \end{subfigure}
    \hfill
    \includegraphics[width=\textwidth]{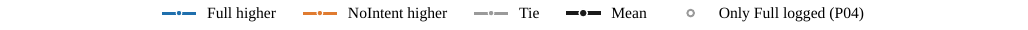}
    \caption{(a) Overall torque alignment counts, (b) torque magnitude, as defined in Eq.~\eqref{eq:torque_magnitude}, and (c) aligned torque, as defined in Eq.~\eqref{eq:torque_difference} for the oncoming car event, where each point represents one participant.  Higher values indicate a higher average torque in alignment with the driver's intent to pass left or right around the oncoming car.
    Lines connect each participant's two conditions.  Panel (a) counts all 12 sessions: P04, drawn hollow, has no \nointent{} trials, and P12, excluded from the subjective analysis, is included.  \withintent{} had more alignment events for 8 participants, tied for 2 and fewer for 1.  Of the 11 participants with both conditions, 7 received higher overall torque under \withintent{} (b) and 8 experienced more intent-aligned torque (c).
    }
    \label{fig:torque_magnitude}
\end{figure*}

\paragraph*{Objective Findings}
We find that \withintent{} outperformed \nointent{} on overall torque alignment, with \withintent{} higher for 8 of the 11 participants who drove both conditions (Fig.~\ref{fig:torque_alignment}; 215 vs.\ 141 events over all 12 sessions, +52\%; P04 has no \nointent{} trials). Collision rates were lower under \withintent{} (2.7\%, 4 of 147 trials) than \nointent{} (7.7\%, 10 of 130 trials); the \nopolicy{} baseline had 1.7\% (2 of 121).  Collisions occurred for four of the ten participants, and one (P08) accounts for 8 of the 16.

\paragraph*{Subjective Findings}
\label{sec:eval3_subjective}
Across the seven Likert dimensions, paired t-tests showed no significant differences between \withintent{} and \nointent{} (see Table~\ref{tab:model_comparison}, all $p > 0.3$). Per-participant overall ratings (Table~\ref{tab:per_participant}) showed high variance in individual differences: five participants rated the intent-aware policy higher and five rated the non-intent-aware policy higher.  Mean overall ratings were nearly identical across conditions (\withintent{} $M = 6.05$, \nointent{} $M = 6.18$).  

Despite the absence of significant rating differences, participants consistently distinguished the two policies in open-ended feedback. Participants described the intent-conditioned policy as collaborative and the no-intent policy as overriding:
\begin{quote}
``[\nointent{}] felt like a captain with too much pushback on the wheel, while [\withintent{}] felt like a copilot helping me out rather than forcing decisions. [\nointent{}] decided which direction I was going[\ldots] [\withintent{}] showed me the way, but if I switched, it would push back only a little.'' (P08)
\end{quote}
\begin{quote}
``[\withintent{}] had the smoothest ride, and less crashes.'' (P05)
\end{quote}
\noindent Nonetheless, not all participants agreed: one preferred how the no-intent policy behaved \textit{after} it corrected to intent:
\begin{quote}
``[\nointent{}] was better. After it corrected to where I was going, I liked it better than [\withintent{}]. I had to fight with it less.'' (P01)
\end{quote}
\noindent For this participant, directional consistency (alignment) was less valued than firm corrective intervention, suggesting that individual differences in preferred autonomy level may explain the alignment-preference relationship.

\paragraph*{Implications}
The results indicate that intent-conditioning yields a measurable objective improvement (alignment, collisions) but a divided subjective response with strong individual differences in policy preference. We also observe that the AI increases collisions, which is likely due to a combination of risk compensation and residual policy errors in close-proximity situations, since participants tended to drive more aggressively when assisted. Disentangling behavioral adaptation from policy-induced risk is an important direction for future work.
This disconnect between objective and subjective findings is taken up in Sec.~\ref{sec:discussion}.

\begin{table}[t]
\caption{Subjective ratings: \initialpolicy{}{} vs.\ \refinedpolicy{}{} ($N = 6$).
  Likert 1--7.
  Paired $t$-tests ($df = 5$).
  This pilot targeted directional trends rather than statistical significance.
  Five out of six participants rated \refinedpolicy{} to be higher on likability.
  }
\label{tab:eval2_subjective}
\centering
\small
\setlength{\tabcolsep}{4pt}
\begin{tabular}{lccrc}
\toprule
Metric & \initialpolicy{} & \refinedpolicy{} & $\Delta$ & $p$ \\
       & $M \pm SE$ & $M \pm SE$ & & \\
\midrule
Likability\,$\uparrow$   & $2.95 \pm .42$ & $3.31 \pm .39$ & $+.36$ & $.108$ \\
Helpfulness\,$\uparrow$  & $2.80 \pm .33$ & $2.85 \pm .20$ & $+.05$ & $.887$ \\
\bottomrule
\end{tabular}
\end{table}

\begin{table}[t]
\caption{Paired comparison of subjective ratings:
  \withintent{} vs.\ \nointent{} ($N = 10$).
  Likert 1--7; Overall Rating 1--10.
  Two-tailed paired $t$-tests ($df = 9$); none reached significance.
  $\uparrow$\,higher is better;
  $\downarrow$\,lower is better (less resistance from AI).
  Participants qualitatively distinguished the two
  policies despite non-significant ratings (Sec.~\ref{sec:eval3_subjective}).}
\label{tab:model_comparison}
\centering
\small
\setlength{\tabcolsep}{4pt}
\begin{tabular}{lccrc}
\toprule
Metric & \withintent{} & \nointent{} & $\Delta$ & $p$ \\
       & $M \pm SE$ & $M \pm SE$ & & \\
\midrule
Directionality\,$\uparrow$     & $4.92 \pm .32$ & $4.65 \pm .29$ & $+.27$ & $.528$ \\
Timing\,$\uparrow$             & $5.10 \pm .17$ & $5.05 \pm .30$ & $+.05$ & $.891$ \\
Adapting Dir.\,$\uparrow$      & $4.90 \pm .36$ & $4.55 \pm .33$ & $+.35$ & $.368$ \\
Fight Dir.\,$\downarrow$       & $3.62 \pm .48$ & $3.62 \pm .27$ & $+.00$ & $1.00$ \\
Safety\,$\uparrow$             & $4.38 \pm .40$ & $4.45 \pm .23$ & $-.07$ & $.870$ \\
Teaming\,$\uparrow$            & $4.58 \pm .39$ & $4.88 \pm .38$ & $-.30$ & $.313$ \\
\midrule
Overall Rating\,$\uparrow$     & $6.05 \pm .40$ & $6.18 \pm .39$ & $-.13$ & $.823$ \\
\bottomrule
\end{tabular}
\end{table}

\begin{table}[t]\caption{Per-Participant Overall Ratings: \withintent{} vs.\ \nointent{}. Overall Rating on a 1-10 scale; values averaged across trials.}
\label{tab:per_participant}
\centering
\begin{tabular}{lccc}
\toprule
Participant & \withintent{} & \nointent{} & $\Delta$ \\
\midrule
P01 & 5.75 & 4.00 & $+1.75$ \\
P02 & 4.75 & 5.75 & $-1.00$ \\
P03 & 7.50 & 4.75 & $+2.75$ \\
P05 & 7.25 & 6.50 & $+0.75$ \\
P06 & 5.50 & 6.50 & $-1.00$ \\
P07 & 5.75 & 8.25 & $-2.50$ \\
P08 & 6.25 & 7.25 & $-1.00$ \\
P09 & 3.50 & 5.75 & $-2.25$ \\
P10 & 7.25 & 7.00 & $+0.25$ \\
P11 & 7.00 & 6.00 & $+1.00$ \\
\midrule
Mean & 6.05 & 6.18 & $-0.13$ \\
\bottomrule
\multicolumn{4}{l}{\footnotesize Participants 4 and 12 excluded (data quality / motion sickness).}\\
\end{tabular}
\end{table}

\subsection{Preliminary On-Vehicle Deployment ($N = 3$)}
To assess transfer beyond the static simulator, we deployed the urban scenario on the GRIP physical test platform~\citep{TRI2024GRIP} with three internal participants (see row 4 of Table~\ref{tab:study_overview}).
Because GRIP's emulated dynamics are sufficiently similar at the low speeds and high-friction conditions to those used in simulation, only minimal parameter adjustments were required to compensate for steering-wheel hardware differences; no retraining was performed. We frame this as a feasibility demonstration of platform transfer rather than a formal evaluation, deferring quantitative on-vehicle evaluation to future work.

The GRIP deployment used the same policy weights and intent space as \textbf{Eval~3} ($\mathcal{Y} = \{\text{left}, \text{right}\}$), with the same reward terms and safety layer (Table~\ref{tab:config_reference}, row~4). Unlike the desktop simulator, GRIP renders the driving scene in VR and communicates the policy's assistive steering torque through the platform's force-feedback hardware, so the driver physically experiences the shared control interaction.  We include three opponent-vehicle scenarios: a vehicle backing out of a driveway, a wrong-way oncoming vehicle, and a roundabout-approach scenario.

\begin{figure*}
    \centering
    \begin{subfigure}[b]{0.99\textwidth}
        \centering
        \includegraphics[width=\textwidth]{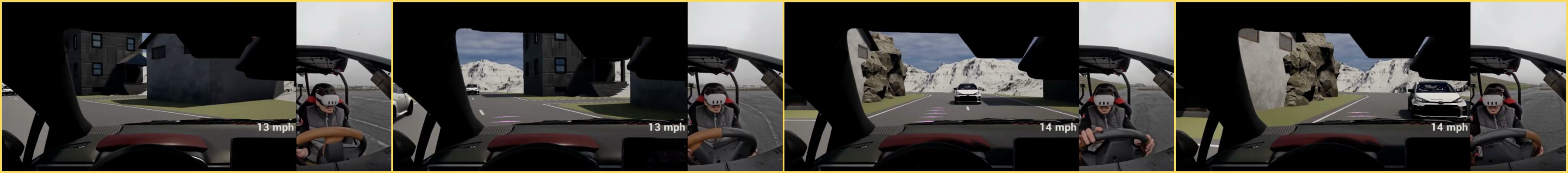}
    \end{subfigure}
    \\[1ex]
    \begin{subfigure}[b]{0.99\textwidth}
        \centering
        \includegraphics[width=\textwidth]{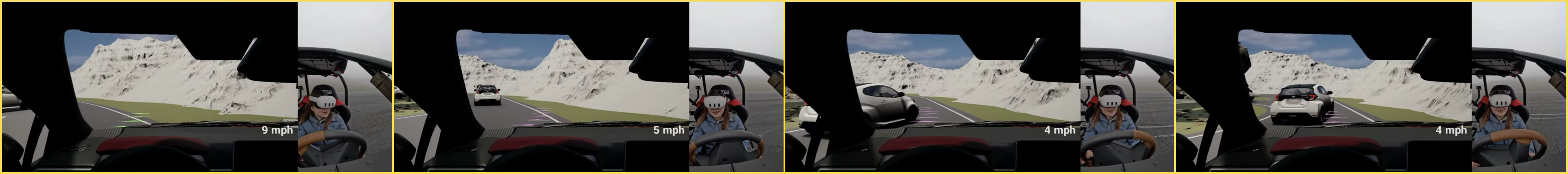}
    \end{subfigure}
    \caption{Two examples from GRIP experiments: an oncoming car scenario where lateral assistance can be observed (top) and a roundabout-approach scenario where longitudinal assistance is observed (bottom).}
    \label{fig:grip_snapshots}
\end{figure*}

In the \emph{oncoming vehicle scenario} (Fig.~\ref{fig:grip_snapshots}, top), a vehicle is seen to be moving toward the ego vehicle down the center lane. The policy infers the driver's lateral avoidance direction and applies assistive steering torque accordingly, guiding the vehicle around the hazard while maintaining speed (13--14 mph). The interaction mirrors the lateral coordination observed in the simulator-based urban evaluation (\textbf{Eval~3}), with the added realism that the driver feels the assistive torque through force-feedback on the wheel.

In the \emph{roundabout-approach scenario} (Fig.~\ref{fig:grip_snapshots}, bottom), the ego vehicle approaches a roundabout entry with a crossing vehicle. In the trial shown, the vehicle decelerates from 9 to 4\,mph as the hazard approaches.  This longitudinal behavior emerges from the task-level collision and progress rewards rather than from an explicit longitudinal intent; the policy learns to slow when the world model predicts an imminent conflict. On GRIP, the driver perceives deceleration through the platform's physical motion cues, allowing the driver to physically feel the effects of longitudinal intervention.  

Together, these scenarios demonstrate that the learned policy transfers to physical hardware and produces qualitatively appropriate assistance without retraining, supporting the feasibility of sim-to-real deployment for shared control.
A video demonstrating the GRIP deployment is available in the
supplementary material (Video~3).

%% file: 8_discussion.tex
\section{Discussion}
\label{sec:discussion}

We observe that, before any refinements, \dreamtoassist{} results perform best on relatively simple, straight sections of the environment (\textbf{Eval~1}), while required iterations via targeted refinements to improve dynamically challenging areas, such as turn sections (\textbf{Eval~2}). The improvements generally helped achieve fuller knowledge of the system's subjective and objective performance in richer environments (\textbf{Eval~3}). Each evaluation informed the construction of a different portion of the proposed workflow in Fig.~\ref{fig:workflow_diagram}. 

This progression followed a particular design and selection methodology.  \textbf{Eval~1} established that the shared-control paradigm works in a constrained setting and identified specific failure mode (over-steering on turns) that motivated targeted refinements. \textbf{Eval~2} confirmed that those refinements improved the interaction in a within-subjects comparison, providing directional evidence with a focused sample. \textbf{Eval~3} deployed the complete framework in a qualitatively different domain under the strongest experimental design (double-blind A/B), yielding the most controlled evidence for intent conditioning. 

\subsection{Intent Conditioning Improves Human-Robot Coordination}
A common result that all three evaluations revealed was that conditioning the assistive policy on intent results in improved physical alignment and improved perception of coordination. In \textbf{Eval~1}, \dreamtoassist{} was directionally consistent with driver intent in turns, and \dreamtoassist{} improved overtake success on straightaways from 80\% to 88\% relative to unassisted driving. Participants rated the policy as understanding their intent significantly above neutral ($M = 5.29$, $p < .001$).  In \textbf{Eval~2}, these results extended to turns. 
In \textbf{Eval~3}, \withintent{} produced 52\% more torque-alignment events across all 12 sessions, and more events for 8 of the 11 participants who drove both conditions.  Collision rates were also lower under \withintent{} (2.7\%) than \nointent{} (7.7\%), though the counts are small and one participant accounts for half of all collisions.

The results validate that extending the RSSM to include an intent predictor on the assistant's latent state, and shaping the assistive reward based on the inferred intent, produces policies that steer \emph{with} the human rather than merely completing the task. We confirm that the value-aligned combination of task completion, intervention sparsity and expert alignment causes this directional consistency. We find that, without the intent predictor, the policy still assists, but bases the assistance on task objectives alone, leading to higher collision rates and qualitatively different interaction dynamics (Sec.~\ref{sec:eval3_subjective}).

\subsection{Bridging the Sim-to-Real Human Gap}
One challenge a system trained with synthetic human partners faces is the gap in simulated and real human behavior (the ``sim-to-real-human gap''). FCP agents produce a diversity of skill levels, propensity for certain discrete behaviors, and noisy behaviors, but can miss certain behaviors such as hesitation, jitter, non-stationary and closed-loop behaviors such as fighting. A world model trained exclusively on these synthetic partners learns to expect smoother, more predictable behavior than what it encounters at deployment. 

\textbf{Eval~1}~$\to$~\textbf{Eval~2} illustrate how this gap manifests and outline how it can be corrected. Qualitative results of \dreamtoassist{} in turns revealed that the intent predictor was not the problem since participants rated intent recognition above neutral even on turns, but was rather due to excessive steering torque that reduced overtake success from 54\% to 27\%. We can thus attribute to the fact that the policy had learned control dynamics calibrated to synthetic partners that behaved differently from real drivers in such a dynamically complex situation.

We therefore introduced targeted refinements, including human-data augmentation (20\% of the replay buffer replaced with real human driving data), FCP output smoothing to suppress unrealistic jitter, and reward restructuring to penalize mode collapse. The result was a directional improvement in \refinedpolicy{} over \withintentorig{} in 59\% of trials, with the largest gains on turns.

The results suggest two general principles.  First, we have evidence that supports that sim-to-real-human gap can be mitigated in simulation-based training by mixing a modest proportion of real human data into the training buffer, which grounds the world model in realistic dynamics, and smoothing synthetic policy outputs reduces the distributional mismatch at the action level. Second, separating intent-inference phenomena from policy-execution phenomena can accelerate correct diagnosis of causes. When objective alignment (enforced through priming human subjects) is high but task performance is poor, the remedy is at the policy level (control quality, reward shaping, safety constraints). When alignment itself is low, the world model or intent predictor could be blameworthy. Our two-level diagnostic is applicable to any architecture that infers latent human objectives and acts on them. We have shown success disentangling such concerns in cases where human evaluation is expensive via a mixture of counterfactual evaluation (offline-replaying logged human control inputs against candidate policies and evaluating), and targeted online human-in-the-loop experiments.

\subsection{Structured Pipeline for Deploying Learned Interactive Policies}
Our proposed policy selection pipeline addresses a practical gap that extends beyond shared driving. Learned policies that perform well against simulated partners may behave poorly with real humans, but exhaustive human evaluation is prohibitive at scale. Currently, no automated metric can fully-capture such qualitative considerations. Hence, we first generate offline metrics to eliminate the bulk of candidates at a negligible per-candidate cost, followed by counterfactual replay to evaluate survivors against recorded human behavior without requiring new participants, and lastly evaluate human-in-the-loop vetting to provide qualitative signal.

We posit that such a structure is relevant to robotics applications. Vision-language-action (VLA) models and language-conditioned manipulation policies face similar scalability problems~\citep{venuto2024code}. The explicit human quality checks fill a gap that we argue will become important when a human-based vetting procedure becomes a practical necessity.

The coupled interface design pipeline (lower row of Fig.~\ref{fig:workflow_diagram}) additionally serves a complementary role where interface-related considerations can be vetted independently of checking policy quality considerations to avoid a dependency of one on the other. We observe that such a separation can be transferrable to other domains where a deployed policy consists of a human-facing component (e.g. LLMs, VLAs).

\subsection{Evaluation Insights: Objective Alignment vs.\ Subjective Preference}
A common finding across all three divergence between objective torque alignment and subjective preference. In \textbf{Eval~3}, intent-conditioned \dreamtoassist{} produced 52\% more torque-alignment events than the no-intent baseline, yet per-participant overall ratings showed no significant difference.  In \textbf{Eval~1}, participants rated the policy as understanding their intent on straightaway, but not on turns, even though the policy was intent-conditioned in both conditions.  This asymmetry suggests that alignment and preference measure different constructs.  Torque alignment captures whether the AI's applied control is directionally consistent with the human's steering input.  The asymmetry suggests that alignment and preference measure different constructs.  Subjective ratings measure retrospective judgments based on specific moments.  Participants notice when the AI opposes them (oversteering in \textbf{Eval~1}, Sec.~\ref{sec:eval1}, and ``captain'' feelings from \textbf{Eval~3}, Sec.~\ref{sec:eval3}), but sometimes do not notice when it supports them as a well-aligned assistance feels like unassisted driving.  On the other hand, torque alignment captures whether the AI's applied control is directionally consistent with the human's steering input aggregated over a trial.  Qualitative feedback from \textbf{Eval~3} supports this interpretation.

This finding parallels observations in the haptic shared-control literature.  \cite{Abbink2018-ei} show that continuous haptic guidance is most effective when it is seamless, i.e., when the driver does not consciously perceive the assistance, and \cite{Mulder2012-fa} show that seamless haptic support improves lane-keeping performance without changing drivers' subjective workload ratings.  Our data extend these findings to a learning-based, intent-conditioned setting: the world model enables the AI to provide directionally consistent assistance that improves an objective alignment metric while remaining perceptually invisible under normal conditions.

\subsection{Generalization Across Domains and Platforms}
We examine the efficacy of \dreamtoassist{} across racing and urban scenarios across multiple fidelities of application - from simulators to GRIP - with longitudinal effects, without structural changes.  We also show that the utility of the RSSM-based intent conditioning helps regardless of domain.  The preliminary results in GRIP demonstrate that the learned policy transfers from simulation to physical platform without retraining with minimal parameter adjustment, and demonstrate viability in a physical driving setting where the human experiences the forces of the interaction, including those provided by longitudinal assistance.

\subsection{Limitations and Future Work}
We note that the three formal evaluations ($N = 24$, $N = 6$, $N = 10$) demonstrate directional effects on alignment and task performance but are underpowered for definitive claims about subjective preference.  A larger sample in \textbf{Eval~3} might reveal some significant trends on preference effects.

Note that in the human studies, the safety filter was applied only to the trajectory visualization and did not modify the assistive torque, ensuring the displayed path remained within safe bounds.  In the four \textbf{Eval~3} sessions with collision logging, both AI conditions produced higher collision rates than the \nopolicy{} baseline (\withintent{}: 2.7\%, \nointent{}: 7.7\%, \nopolicy{}: 1.7\%).  The collision rate differences therefore reflect the learned policy's behavior and the driver's response to it, not the safety filter's intervention. Intent conditioning reduces collisions relative to no-intent AI, but the increase relative to unassisted driving warrants investigation.  One hypothesis is that participants may drive more aggressively when assisted or rely on the AI to manage proximity, expecting the system to compensate for risk.  Understanding how actual policy-induced collisions arise from behavioral adaptation is a promising future direction.

During training, our framework assumes a fixed intent label for each FCP episode. At deployment, the intent predictor updates at every timestep and can track changes in driver behavior; however, because transitions were not explicitly modeled during training, the predictor may lag during rapid intent switches (this lag was studied in~\cite{decastro2025dreaming}). To further improve responsiveness, incorporating explicit intent-transition modeling is a direction for future work.

Preliminary testing on GRIP with three internal participants confirmed that the integration pipeline generalizes to physical hardware, but surfaced new challenges. Participants reported that the early longitudinal policy felt more intrusive on the motion platform than lateral assistance, likely because the driver physically experiences deceleration forces that are absent in the desktop simulator. Tuning the haptic channel for longitudinal behaviors and conducting a formal on-vehicle evaluation remain natural future extensions.

To address the limitations, three promising directions emerge: exploring language-based assistance, where the policy plans with language and physical assistive actions; training with awareness of misalignment between the human's intent and the robot's actions, rather than maximizing alignment alone; and exploring larger-scale studies with real vehicles to validate the results in physical settings.

%% file: 7_conclusions.tex
\section{Conclusion}
We have presented \dreamtoassist{}, an intent-conditioned world model for shared control that infers a human partner's discrete objectives from observed behavior and conditions its assistive policy accordingly. Extending the recurrent state-space model with an intent predictor and a value-aligned reward structure, the framework produces policies that steer with the human's objective rather than merely completing the task.

We report results of three human-subjects evaluations across two qualitatively distinct domains: high-speed racing and surprise-rich urban driving, and demonstrate that intent conditioning yields measurable improvements in human-robot coordination. In the double-blind urban evaluation, the intent conditioning produced 52\% more torque-alignment events and fewer collisions (2.7\% vs.\ 7.7\% of trials) than a non-intent-conditioned baseline, and fewer collisions in the sessions where collisions were logged, and showed transferability to a physical motion platform with no retraining. The racing evaluations established initial feasibility, highlighted areas for refinement to address the sim-to-real-human gap in policies trained with synthetic partners.

Beyond the algorithmic validation, we also contribute a structured deployment pipeline for deploying policies in human-centric domains, that spans automated policy screening from counterfactual log replay to human-in-the-loop testing. A coupled interface-design process ensures that the communication channel between the policy and the driver is resolved before policy quality is formally evaluated. We believe this pipeline generalizes to any domain where learned policies must be vetted before deployment with human partners.

There are several areas for further exploration. The fixed-intent assumption limits applicability to scenarios where the human's objective is stable within an episode; supporting online intent transitions is a natural extension. The individual differences in preferred autonomy level observed across participants motivate adaptive policies that modulate intervention style per driver. And while the GRIP deployment demonstrates feasibility of platform transfer, a powered on-vehicle study is needed to substantiate claims about real-world deployment.

%% file: A_model_architecture.tex
 
\section{Model Architecture}
\label{app:model_architecture}
 
This appendix details the neural network architecture and training
hyperparameters for the \dreamtoassist{} world model described in
Sec.~\ref{sec:approach}.  Both the human ($\kappa = H$) and
assistive ($\kappa = A$) agents share the same architectural template
but maintain separate parameters $W_\phi^H$ and $W_\phi^A$.
The architecture is shared across domains; domain-specific differences
are limited to the observation/action spaces, reward configuration
(Appendix~\ref{sec:policy_interface_design}), and replay buffer composition.
 
\subsection{RSSM Components}
 
The recurrent state-space model comprises the components listed in
Table~\ref{tab:rssm_architecture}.  All modules operate on the
concatenated hidden state $s_t^\kappa = \{h_t^\kappa, z_t^\kappa\}$,
where $h_t^\kappa \in \mathbb{R}^{512}$ is the deterministic
recurrent state and $z_t^\kappa$ is the stochastic discrete
representation.
 
\begin{table*}[h]
\centering
\small
\caption{RSSM component architecture.  All MLP layers use SiLU
  activations and LayerNorm.  ``Cat.'' denotes a categorical
  distribution parameterized by logits.}
\label{tab:rssm_architecture}
\begin{tabular}{lll}
\toprule
\textbf{Component} & \textbf{Architecture} & \textbf{Output} \\
\midrule
Encoder $q_\phi^\kappa$
  & MLP: 2 layers $\times$ 512 units
  & Input to posterior \\
Recurrent model $f_\phi^\kappa$
  & GRU w/ LayerNorm, 512 hidden
  & $h_t^\kappa \in \mathbb{R}^{512}$ \\
Posterior $q_\phi^\kappa(z_t | h_t, x_t)$
  & MLP: 1 layer $\times$ 512 units
  & $z_t^\kappa$ (Cat.\ $32 \times 32$) \\
Prior $p_\phi^\kappa(\hat{z}_t | h_t)$
  & MLP: 1 layer $\times$ 512 units
  & $\hat{z}_t^\kappa$ (Cat.\ $32 \times 32$) \\
Decoder $p_\phi^\kappa(\hat{x}_t | s_t)$
  & MLP: 2 layers $\times$ 512 units
  & $\hat{x}_t^\kappa$ (symlog MSE) \\
Reward head $p_\phi^\kappa(\hat{r}_t | s_t)$
  & MLP: 2 layers $\times$ 512 units
  & symlog disc. \\
Continue head $p_\phi^\kappa(\hat{c}_t | s_t)$
  & MLP: 2 layers $\times$ 512 units
  & Bernoulli \\
Intent head $p_\phi^A(\hat{y}_t | s_t^A)$
  & MLP: 2 layers $\times$ 512 units
  & $|\mathcal{Y}|$ classes (one-hot) \\
\bottomrule
\end{tabular}
\end{table*}
 
The discrete representation $z_t^\kappa$ follows the DreamerV3
parameterization: a product of 32 categorical variables, each with
32 classes, yielding a $32 \times 32 = 1024$-dimensional one-hot
vector that is flattened and concatenated with $h_t^\kappa$ to form
$s_t^\kappa \in \mathbb{R}^{1536}$.  Unimix regularization
($\epsilon = 0.01$) prevents mode collapse in the categorical
distributions.
 
\subsection{Observation and Action Spaces}
 
Both domains use proprioceptive (state-based) observations rather
than pixel inputs.  The encoder and decoder operate on MLP keys only
(no CNN).  Table~\ref{tab:obs_action_spaces} summarizes the
observation and action spaces.
 
\begin{table*}[h]
\centering
\small
\caption{Observation and action spaces by domain.  All observations
  are expressed in the ego vehicle's body frame.}
\label{tab:obs_action_spaces}
\begin{tabular}{lcc}
\toprule
& \textbf{Racing (CARLA)} & \textbf{Urban / GRIP} \\
\midrule
Num.\ agents in env & 2 & up to 10 \\
Max.\ opponents in obs & 1 & 3 \\
Action repeat & 2 & 2 \\
Episode time limit & 1000 steps & 1000 steps \\
\midrule
\multicolumn{3}{l}{\textit{Observation components (ego-centric)}} \\
\quad Ego state & \multicolumn{2}{c}{position, velocity, heading (rel.\ to track)} \\
\quad Opponent state(s) & 1 opponent & up to 3 opponents \\
\quad Road geometry & track centerline, edges & road edges, intersections \\
\midrule
Human action dim & \multicolumn{2}{c}{3 (steering, throttle, brake)} \\
AI action dim & \multicolumn{2}{c}{3 (steering torque, throttle, brake)} \\
\bottomrule
\end{tabular}
\end{table*}
 
\subsection{Actor-Critic Architecture}
 
The actor and critic networks (Eq.~\eqref{eq:ac}) are implemented as:
 
\begin{itemize}
\item \textbf{Actor} $\pi_\psi^\kappa(a_t^\kappa | s_t)$:
  MLP with 2 hidden layers of 512 units (SiLU, LayerNorm),
  outputting the mean and log-std of a normal distribution over
  continuous actions.  Action unimix ratio $\epsilon = 0.01$.
  Minimum std $= 0.1$, maximum std $= 1.0$.
 
\item \textbf{Critic} $v_\psi^\kappa(s_t)$:
  MLP with 2 hidden layers of 512 units (SiLU, LayerNorm),
  outputting a symlog-discretized scalar value estimate.
  Trained with $\lambda$-returns
  ($\lambda = 0.95$, $\gamma = 0.997$) and slow-target
  regularization (EMA fraction $= 0.02$).
\end{itemize}
 
\subsection{Training Configuration}
 
Table~\ref{tab:training_config} lists the key training
hyperparameters.  Values are shared across domains unless otherwise
noted.
 
\begin{table}[h]
\centering
\small
\caption{Training hyperparameters.  All values are shared between
  the racing and urban configurations unless marked.}
\label{tab:training_config}
\begin{tabular}{lc}
\toprule
\textbf{Hyperparameter} & \textbf{Value} \\
\midrule
\multicolumn{2}{l}{\textit{World model}} \\
Batch size & 16 \\
Sequence length & 64 \\
Train ratio & 512 \\
Learning rate & $1 \times 10^{-4}$ \\
Optimizer & Adam ($\epsilon = 10^{-8}$) \\
Gradient clip & 1000 \\
KL balancing ($\beta_{\text{dyn}} / \beta_{\text{rep}}$) & 0.5 / 0.1 \\
Free nats & 1.0 \\
Reward loss scale & 1.0 \\
Continue loss scale & 1.0 \\
Intent loss scale & 0.1 \\
Training steps & $2 \times 10^6$ \\
\midrule
\multicolumn{2}{l}{\textit{Actor-critic}} \\
Actor learning rate & $3 \times 10^{-5}$ \\
Critic learning rate & $3 \times 10^{-5}$ \\
Actor/critic optimizer & Adam ($\epsilon = 10^{-5}$) \\
Actor/critic gradient clip & 100 \\
Imagination horizon & 15 steps \\
Discount $\gamma$ & 0.997 \\
$\lambda$-return $\lambda$ & 0.95 \\
Actor distribution & Normal \\
Actor entropy bonus & $3 \times 10^{-4}$ \\
Slow target EMA fraction & 0.02 \\
\midrule
\multicolumn{2}{l}{\textit{Domain-specific}} \\
Replay buffer (racing) & $10{,}000$ episodes \\
Replay buffer (urban) & $1{,}000{,}000$ transitions \\
Human data ratio (Evals 2--3) & 20\% of replay buffer \\
Intent classes $|\mathcal{Y}|$ & 2 (all domains) \\
\midrule
\multicolumn{2}{l}{\textit{Safety layer (Sec.~\ref{sec:safety_layer})}} \\
CBF $\alpha$ & 2.0 \\
Inner collision radius $\tau_{\text{inner}}$ & 3.0\,m \\
Outer collision radius $\tau_{\text{outer}}$ & 6.0\,m \\
\bottomrule
\end{tabular}
\end{table}
 
\subsection{FCP Human Population}
 
The fictitious co-play procedure (Sec.~\ref{sec:pop_human_model}) generates
separate sub-populations for each intent $y \in \mathcal{Y}$.
In the racing domain, two human policies are trained---one for
\emph{pass} and one for \emph{stay behind}---and checkpoints are
saved at regular intervals during training.  Each checkpoint
$k \in [1, K]$ represents a policy trained to $k\%$ completion,
with the final checkpoint $k = K$ approximating the optimal policy
$\pi_y^*$.  The full population
$\Pi_y^H = \{\pi_{y,k}^H\}_k \cup \{\pi_y^*\}$ spans the skill
spectrum from novice to expert for each intent.
 
In the urban domain, three human sub-populations are trained:
\emph{left}, \emph{right}, and \emph{fast} (speed-seeking).
Population sampling is biased to reflect the expected frequency
of each behavior (left: 46.2\%, fast: 34.0\%, right: 19.8\%).
During AI training, 100\% of episodes use population play
(i.e., the frozen human agent is always drawn from the FCP
population).
 
\subsection{Reward Configuration Summary}
 
Table~\ref{tab:reward_config_summary} summarizes the key reward
scalars that differ between domains and training roles (human vs.\
AI).  Full reward definitions are given in
Appendix~\ref{sec:policy_interface_design}.
 
\begin{table}[h]
\centering
\small
\caption{Key reward scalars by domain and training role.  ``Human''
  denotes training of FCP sub-populations; ``AI'' denotes training
  of the assistive policy.}
\label{tab:reward_config_summary}
\begin{tabular}{lccc}
\toprule
\textbf{Parameter} & \textbf{Racing} & \textbf{Urban} \\
\midrule
\multicolumn{3}{l}{\textit{Collision penalties}} \\
\quad Inner collision & $-5{,}000$ & $-10{,}000$ \\
\quad Outer collision & --- & $-10{,}000$ \\
\quad Boundary violation & $-100$ & $-10{,}000$ \\
\midrule
\multicolumn{3}{l}{\textit{Intent rewards (human training)}} \\
\quad Pass/left reward scale & 10 & $\alpha_{\text{lat}}^+$ \\
\quad Stay/right reward scale & $-10$ & $\alpha_{\text{lat}}^-$ \\
\quad Left/right bias penalty & --- & 1{,}000 \\
\quad Bias distance threshold & --- & 8.0\,m \\
\midrule
\multicolumn{3}{l}{\textit{AI-specific}} \\
\quad Intervention penalty $\|a_t^A\|_2$ & 5{,}000 & 5{,}000 \\
\quad Pass/progress reward scale & 0 & 0 \\
\quad Speed bonus multiplier & --- & 10.0 \\
\bottomrule
\end{tabular}
\end{table}

%% file: B_interface_implementation.tex
\section{Policy and Interface Design}
\label{sec:policy_interface_design}

Beyond the initial algorithm and trained policy, deploying a system
that interacts with humans in a tight loop requires careful shaping
of both the deployed policy and the interface through which it
communicates with the driver.
 
\subsection{Policy Shaping}
 
The reward terms below instantiate the general framework of
Sec.~\ref{sec:approach}.  The task reward $r_t^{task}$ \eqref{eq:task_specific_rews}
comprises collision penalties Eq.~\eqref{eq:collision}, boundary
penalties \eqref{eq:bound}--\eqref{eq:soft}, an action smoothness
penalty \eqref{eq:mu}--\eqref{eq:smooth}, and a speed/progress
bonus \eqref{eq:speed}--\eqref{eq:tile}.  These are shared across
domains with adapted parameters.  The human-objective reward $r_t^y$,
conditioned on the active intent $y \in \mathcal{Y}$, is
domain-specific: pass/stay rewards
\eqref{eq:rew_pass_stay}--\eqref{eq:rew_total_pass_prog} for
racing ($\mathcal{Y} = \{\text{overtake}, \text{stay\_behind}\}$),
left/right rewards \eqref{eq:leftright} for urban driving
($\mathcal{Y} = \{\text{left}, \text{right}\}$), and stop/go rewards
\eqref{eq:stopgo} for the GRIP deployment
($\mathcal{Y} = \{\text{left}, \text{right}\} \times
\{\text{go}, \text{slow}, \text{stop}\}$).
Table~\ref{tab:reward_mapping} summarizes which terms are active per
domain.  We now describe each in detail.

\begin{table}[h]
\caption{Mapping of reward terms to the general framework
  (Eqs.~\eqref{eq:task_specific_rews} and \eqref{eq:ai_reward}) and to each domain.
  \cmark\ = active; \xmark\ = not used.}
\label{tab:reward_mapping}
\centering
\small
\setlength{\tabcolsep}{3.5pt}
\resizebox{\columnwidth}{!}{%
\begin{tabular}{l c c c c}
\toprule
\textbf{Reward term} & \textbf{Type} & \textbf{Racing} & \textbf{Urban} & \textbf{GRIP} \\
\midrule
Collision Eq.~\eqref{eq:collision}
  & $r^{task}$ & \cmark & \cmark & \cmark \\
Boundary Eq.~\eqref{eq:bound}--\eqref{eq:soft}
  & $r^{task}$ & \cmark & \cmark & \cmark \\
Smoothness Eq.~\eqref{eq:mu}--\eqref{eq:smooth}
  & $r^{task}$ & \cmark & \cmark & \cmark \\
Speed / progress Eq.~\eqref{eq:speed}--\eqref{eq:tile}
  & $r^{task}$ & \cmark & \cmark & \cmark \\
Finish-line bonus
  & $r^{task}$ & \cmark & \xmark & \xmark \\
\midrule
Pass / stay Eq.~\eqref{eq:rew_pass_stay}--\eqref{eq:rew_total_pass_prog}
  & $r^y$ & \cmark & \xmark & \xmark \\
Left / right \eqref{eq:leftright}
  & $r^y$ & \xmark & \cmark & \cmark \\
\midrule
Intervention sparsity Eq.~\eqref{eq:ai_reward}
  & $r^{interv}$ & \cmark & \cmark & \cmark \\
Expert alignment Eq.~\eqref{eq:ai_reward}
  & --- & \cmark & \cmark & \cmark \\
\bottomrule
\end{tabular}%
}
\end{table}

\subsubsection{Pass-Stay Rewards}
Mode collapse occurs when the policy converges to always passing or always staying behind, regardless of the scenario's intended behavior.  This was observed in the characterization study.  The reward restructuring introduced in \textbf{Eval~2} addresses this by supporting both behaviors and penalizing deviation from the intended one.  In the initial configuration, a dense reward was given when ego was ahead of the opponent vehicle within a certain pass range window.  In the refinement study, we adopt:
{\small
\begin{align}
\label{eq:rew_pass_stay}
    r_{\text{pass}} &= 
    \begin{cases} 
    \alpha_{\text{pass}}^{+} \mathbf{1}\left[ d_{\text{low}}^{+} < d < d_{\text{high}}^{+} \right] & \text{if intent} = \texttt{overtake} \\[6pt]
    \alpha_{\text{pass}}^{-} \mathbf{1}\left[ d_{\text{low}}^{-} < d < d_{\text{high}}^{-} \right] & \text{if intent} = \texttt{stay\_behind}
    \end{cases}
\end{align}
}
\begin{align}
\label{eq:rew_prog_correction}
    r_{\text{progress}} = 
    \begin{cases} 
    \beta^{+} v_{\text{ego}} & \text{if intent} = \texttt{overtake} \\[6pt]
    \beta^{-} v_{\text{ego}} & \text{if intent} = \texttt{stay\_behind}
    \end{cases}
\end{align}
\begin{align}
\label{eq:rew_total_pass_prog}
    r_{\text{total}} = \min\left(r_{\text{cap}}, r_{\text{pass}} + r_{\text{progress}} \right)
\end{align}
where $\alpha_{\text{pass}}^{+}, \alpha_{\text{pass}}^{-}$ are pass reward scalars, 
$d_{\text{low}}^{\pm}, d_{\text{high}}^{\pm}$ are pass low/high bounds per intent,
$\beta^{+}, \beta^{-}$ are progress reward scalars for overtake and stay,
$\mathbf{1}[\cdot]$ is the indicator function.

\subsubsection{Left-Right Rewards}
\label{sec:leftright_rewards}
 
In the urban domain, the opponent vehicle approaches from the front
and the driver must choose to avoid it on the \emph{left} or
\emph{right}.  The intent-conditioned reward penalizes lateral
positions inconsistent with the inferred intent.  Let
$\ell_{\text{opp}}$ denote the lateral position of the opponent in
the ego's body frame, and let $y_{\text{lat}} \in \{-1,+1\}$ encode
the inferred lateral intent (left or right).  When the ego is within
a proximity threshold $d_{\text{lat}}$ of the opponent, we apply:
\begin{align}
\label{eq:leftright}
r_{\text{lateral}} =&
\begin{cases}
  \alpha_{\text{lat}}^{+}
    & \text{if } \text{sign}(\ell_{\text{ego}} - \ell_{\text{opp}})
      = y_{\text{lat}} \\[4pt]
  -\alpha_{\text{lat}}^{-}
    & \text{if } \text{sign}(\ell_{\text{ego}} - \ell_{\text{opp}})
      \neq y_{\text{lat}}
\end{cases},\\ \nonumber
&\quad \text{if } \|\mathbf{x}_{\text{ego}} - \mathbf{x}_{\text{opp}}\|
  < d_{\text{lat}}
\end{align}
where $\alpha_{\text{lat}}^{+}$ and $\alpha_{\text{lat}}^{-}$ are
positive and negative reward scalars, respectively.  Outside the
proximity threshold, $r_{\text{lateral}} = 0$.  This reward
encourages the policy to position the ego on the correct side of
the opponent relative to the inferred intent, penalizing
inconsistent lateral positioning.

\subsubsection{Stop-Go Rewards}
\label{sec:stopgo_rewards}
 
The GRIP deployment extends the urban intent space with longitudinal
intents that we term \emph{go} (maintain or increase speed),
\emph{slow} (decelerate but continue), and \emph{stop} (come to a
halt).  As in manipulation and locomotion RL, where careful reward
shaping is needed to elicit distinct discrete
behaviors~\citep{ng1999policy}, we find that sparse,
intent-specific reward terms are essential to prevent mode collapse
across longitudinal modes.  The longitudinal reward modulates the
speed and collision terms based on the inferred longitudinal intent:
\begin{equation}
\label{eq:stopgo}
r_{\text{long}} =
\begin{cases}
  \beta_{\text{go}}^{+} \, v_{\text{ego}}
    & \text{if intent} = \texttt{go} \\[4pt]
  -\beta_{\text{slow}} \,
    \mathbf{1}\!\left[\, v_{\text{ego}} > v_{\text{thresh}} \,\right]
    & \text{if intent} = \texttt{slow} \\[4pt]
  -\beta_{\text{stop}} \, v_{\text{ego}}
    & \text{if intent} = \texttt{stop}
\end{cases}
\end{equation}
where $v_{\text{ego}}$ is the ego's forward speed,
$v_{\text{thresh}}$ is a speed threshold for the \emph{slow} intent,
and $\beta_{\text{go}}^{+}$, $\beta_{\text{slow}}$,
$\beta_{\text{stop}}$ are scalar weights.  Under the \emph{stop}
intent, the default speed/progress bonus (Eq.~\eqref{eq:speed}) is
additionally suppressed to avoid conflicting objectives.  Stopping
behaviors also modify the collision penalty thresholds to account
for the reduced stopping distance at lower speeds.

\subsubsection{Collision / Bound Rewards}
Proximity between the ego and each opponent vehicle is evaluated using a three-circle approximation, where the distance of the ego's front, center, and rear points to those of the opponent are calculated, and the minimum of each $d_{\min}$ is taken.  We then define two collision zones:
\begin{equation}
\label{eq:collision}
    p_{\text{collision}} =
    \begin{cases}
    C_{\text{inner}} & \text{if } d_{\min} < \tau_{\text{inner}} \\
    C_{\text{outer}} & \text{if } \tau_{\text{inner}} \leq d_{\min} < \tau_{\text{outer}} \\
    0 & \text{otherwise}
    \end{cases},
\end{equation}
where $\tau_{\text{inner}} = 3.0\text{m}$, $\tau_{\text{outer}} = 6.0\text{m}$ define the inner and outer collision zones, and $C_{\text{inner}} = C_{\text{outer}} = 10,000$.  An inner collision can optionally terminate the episode.  We terminate in the racing domain to encourage safety, but do not terminate in the urban domain as we found that terminating discouraged exploration.

If the ego's position, when projected into the local track frame, falls outside the left and right track edges at the nearest tile, a large penalty is incurred:
\begin{equation}
\label{eq:bound}
    p_{\text{bound}} =
    \begin{cases}
    B_{\text{hard}} & \text{if ego is outside track edges} \\
    0 & \text{otherwise}
    \end{cases}
\end{equation}
where $B_{\text{hard}} = 10,000$.

As the ego approaches the track edge, a penalty ramps up linearly:
\begin{equation}
\label{eq:soft}
    p_{\text{soft}} = B_{\text{soft}} \left(1 - \frac{\min(d_L, d_R)}{\tau_{\text{soft}}}\right) \quad \text{if } \min(d_L, d_R) < \tau_{\text{soft}}
\end{equation}
where $d_L$, $d_R$ are the distances to the left and right track boundaries, $B_{\text{soft}} = 1000$, and $\tau_{\text{soft}} = 0.5\text{m}$.

\subsubsection{Action Smoothness Penalty}
To discourage jittery human policy steering and AI interventions, a penalty is applied to the magnitude of the frame-to-frame change in steering angle. The penalty weight grows with training progress through a scheduled multiplier:
\begin{equation}
\label{eq:mu}
    \mu(t) = \min\left(\mu_{\max}, \lambda t\right)
\end{equation}
\begin{equation}
\label{eq:smooth}
    p_{\text{smooth}} = \mu(t)\left\|\delta_{\text{steer}}^{(t)} - \delta_{\text{steer}}^{(t-1)}\right\|
\end{equation}
$\mu_{\max} = 20$, $\lambda = 0.0002$. Early in training the penalty is negligible, allowing the policy to explore freely; later it grows to enforce smooth steering.

\subsubsection{Speed and Progress Bonus}
When the ego is not within the outer collision zone, a reward proportional to the track-relative forward velocity encourages the ego to maintain speed:
\begin{equation}
\label{eq:speed}
    b_{\text{speed}} = \beta v_{\text{ego}}^{\text{track}} \quad \text{if } d_{\min} \geq \tau_{\text{outer}}
\end{equation}
where $\beta = 1.0$ by default. The bonus is suppressed near the opponent to avoid rewarding aggressive behavior.

The map is subdivided into a set of tiles, each representing a segment of road surface the agent must traverse to aggregate a reward bonus.  In racing, there is an added incentive to achieve higher bonus if the tile has not been visited yet.
\begin{equation}
\label{eq:tile}
    r_{\text{tile}}^{(i)} = \phi \frac{1000}{|\mathcal{T}|}
\end{equation}
where $|\mathcal{T}|$ is the total number of track tiles and $\phi$ is a dampening factor:
\begin{equation}
    \phi =
    \begin{cases}
    1 - \dfrac{n_{\text{prev}}}{N_{\text{agents}}} & \text{if } \texttt{is\_racing} = \texttt{True} \\
    1 & \text{otherwise}
    \end{cases}
\end{equation}
where $n_{\text{prev}}$ is the number of other cars that have already visited the tile. When $N_{\text{agents}} = 2$ and the ego is the second car to reach a tile, $\phi = 0.5$.

\subsubsection{Smoothing Human / AI Policy Outputs}
Raw action outputs from a policy can exhibit high-frequency
oscillations that are physically unrealistic.  When a fictitious
human policy is jittery, it can cause the assistant policy to learn
behaviors adapted to such artifacts rather than to smooth human
driving.  We therefore smooth the fictitious human policy's raw
actions at each timestep using a simple causal moving-average filter
over the most recent $N$ actions.  We select a window length of
$N = 10$ steps to approximate realistic steering dynamics.  A
similar smoothing is applied to the output of the AI policy before
feeding it to the physical wheel and the visual trajectory-line
generator.  This smoothing was introduced in \textbf{Eval~2} and
used in all subsequent evaluations.

\subsection{Interface Design Considerations}
\label{sec:interface}

We define the \emph{human-machine interface} (HMI) as all channels through which \dreamtoassist{} communicates its actions to the driver and influences the driver's control inputs, including haptic steering torque, visual trajectory cues, and auditory feedback.
The interface itself was refined via the iterative process described
in Sec.~\ref{sec:deploy}.  The main feedback mechanisms explored
were visual trajectory cues and haptic steering feedback.
 
We explored several variations to optimize explainability and
transparency of the policy's actions while balancing the driver's
cognitive load.  These included showing the policy's steering
actions via trajectory lines, audio utterances, displaying the
system's inferred intent (to overtake or not), and varying color
combinations and line lengths.  The full set of design variables is
listed in Table~\ref{tab:HF_factors}.  Below we describe each
category.

\begin{table*}[t]
\centering
\footnotesize
\caption{\label{tab:HF_factors}Design variables and human factors
  considered in \dreamtoassist{}, organized by category.
  Domain and platform indicate where each factor was actively
  applied.  Factors marked \emph{(excluded)} were considered but
  rejected during piloting.}
\setlength{\tabcolsep}{5pt}
\renewcommand{\arraystretch}{1.25}
\resizebox{\linewidth}{!}{%
\begin{tabular}{P{3.4cm} P{8.8cm} P{2.2cm} P{2.0cm}}
\toprule
\textbf{Name} & \textbf{Description} & \textbf{Domain} & \textbf{Platform} \\
\midrule
\multicolumn{4}{l}{\cellcolor{sectionbg}\textbf{Policy Training}} \\
\midrule
Real human data augmentation &
  20\% of replay buffer replaced with relabeled real human driving
  data to reduce sim-to-real gap. &
  \tagRacing\ \tagUrban & \tagSim \\
FCP population diversity &
  Synthetic humans sampled at varying training checkpoints
  (suboptimal to optimal) to expose the assistant to diverse
  behaviors. &
  \tagRacing\ \tagUrban & \tagSim \\
Action smoothing &
  Causal moving-average (window $= 10$) applied to synthetic human
  and AI outputs to suppress physically unrealistic jitter. &
  \tagRacing\ \tagUrban & \tagSim\ \tagGRIP \\
Intent class re-balancing &
  Training reweights intent classes to prevent mode collapse toward
  a dominant intent. &
  \tagUrban & \tagSim \\
Pass/stay reward restructuring &
  Bounded per-intent indicator reward plus speed progress term,
  capped to prevent single-mode collapse
  (Eq.~\eqref{eq:rew_pass_stay}--\eqref{eq:rew_total_pass_prog}). &
  \tagRacing & \tagSim \\
Left/right rewards &
  Directional reward shaping for lateral urban intents
  (Eq.~\eqref{eq:leftright}). &
  \tagUrban & \tagSim\ \tagGRIP \\
Stop/go rewards &
  Longitudinal reward shaping for speed intents
  (Eq.~\eqref{eq:stopgo}). &
  \tagUrban & \tagGRIP \\
Intervention sparsity penalty &
  $L_2$-norm on AI action magnitude, forming a mixed $L_1$--$L_2$
  norm over time to encourage sparse, targeted interventions. &
  \tagRacing\ \tagUrban & \tagSim\ \tagGRIP \\
\midrule
\multicolumn{4}{l}{\cellcolor{sectionbg}\textbf{Safety Layer}} \\
\midrule
CBF safety filter &
  First-order control barrier function constraining the displayed trajectory (not the assistive torque) to satisfy lane and obstacle safety;
  $\delta a$ solved analytically (Eq.~\eqref{eq:cbf}). &
  \tagUrban & \tagSim\ \tagGRIP \\
\midrule
\multicolumn{4}{l}{\cellcolor{sectionbg}\textbf{Policy Selection Pipeline}} \\
\midrule
FCP evaluation statistics &
  Stage~1: threshold screening on training-time metrics (task
  progress, collision rate, intent accuracy, intervention
  sparsity). &
  \tagRacing\ \tagUrban & \tagSim \\
Counterfactual log-replay &
  Stage~2: policies scored on replayed human logs via collision
  propensity and track-orthogonality of control inputs. &
  \tagRacing\ \tagUrban & \tagSim \\
Human-in-the-loop policy gate &
  Stage~3: internal pilots rate candidates pass/fail on predefined
  criteria before external study deployment. &
  \tagRacing\ \tagUrban & \tagSim \\
\midrule
\multicolumn{4}{l}{\cellcolor{sectionbg}\textbf{Interface \& HMI Design}} \\
\midrule
Trajectory line presence \& length &
  RSSM future path overlaid on scene; length scaled by opponent
  proximity (racing) or assistance magnitude (urban). &
  \tagRacing\ \tagUrban & \tagSim\ \tagGRIP \\
Trajectory line color coding &
  Green/red encodes acceleration vs.\ deceleration; chosen over bar
  graphs and separate throttle/brake displays. &
  \tagRacing\ \tagUrban & \tagSim\ \tagGRIP \\
Trajectory line temporal filtering &
  Mean filter over recent rollouts smooths displayed line to reduce
  perceptual jitter. &
  \tagRacing\ \tagUrban & \tagSim\ \tagGRIP \\
Haptic force feedback (steering) &
  Wheel torque blended from human and policy setpoints; gain and
  resistance tuned iteratively to communicate without overwhelming
  driver input. &
  \tagRacing\ \tagUrban & \tagSim\ \tagGRIP \\
Longitudinal feedback (engine + line) &
  Throttle/brake assistance communicated via engine noise and line
  color, as haptic pedal feedback was impractical. &
  \tagRacing\ \tagUrban & \tagSim \\
Audio utterance cues \emph{(excl.)} &
  Rule-triggered speech cues evaluated as trajectory line
  complement; excluded for latency, sparsity, and cognitive load. &
  \tagRacing & \tagSim \\
Bar graph display \emph{(excl.)} &
  Brake/throttle/steering bar graphs considered; excluded for
  increased cognitive load. &
  \tagRacing & \tagSim \\
\midrule
\multicolumn{4}{l}{\cellcolor{sectionbg}\textbf{Physical Platform Transfer}} \\
\midrule
Steering hardware calibration &
  Minimal parameter adjustments between simulator (G29/Fanatec) and
  GRIP to compensate for hardware differences. &
  \tagUrban & \tagGRIP \\
\bottomrule
\end{tabular}%
}
\end{table*}

\subsubsection{Trajectory Visualization}
\label{sec:traj_viz}

A trajectory visualization is created by rolling out the RSSM world
model forward in time, capped at 7 rollout steps (approximately two
seconds of future travel at the simulation timestep).  The
resulting path is projected onto the simulated world to emulate a
heads-up display.  To ensure visual consistency, the displayed line
is filtered using a mean filter averaged over a window of
trajectories generated across several recent timesteps.  We
performed several design iterations, exploring when the line should
appear, its color, type, and length.  The final design encodes
whether the policy is applying acceleration (green) vs.\
deceleration (red) via line color.  In racing, the line length is
fixed at the full rollout horizon.  In the urban setting, line
length is additionally scaled by the magnitude of the assistance
(up to the rollout cap), such that weaker assistance produces a
shorter displayed line.

\subsubsection{Haptic Feedback}

Force feedback gain (the mapping from policy torque to physical
wheel torque), steering resistance (the baseline centering force
felt by the driver), and torque blending ratio (the relative weight
of human vs.\ policy setpoints) were tuned through iterative pilot
sessions to ensure the haptic channel communicated assistance
without overwhelming the driver's control input.  Due to the difficulty in
applying haptics to the pedals (the throttle/brake channel is
uncoupled; see Sec.~\ref{sec:sim_infrastructure}), longitudinal
assistance was communicated via engine noise and trajectory line
color.

\subsubsection{Additional Modalities}
We further experimented with a fixed taxonomy of audio utterances
triggered via rules applied to the policy outputs (e.g., saying
``Adding speed and nudging you to the left'' when the policy outputs
noticeable acceleration and left-directed steering).  Experiments
were conducted to determine if this modality could complement or
replace visual trajectory lines.  Because participants rated the
utterances as too delayed, too sparse, and difficult to comprehend
while driving, these audio cues were excluded from the final design.

\subsubsection{Iterative Piloting \& Main Takeaways}
 
Each time a new version was ready for testing, we conducted piloting
sessions to gather human feedback.  These pilots included both
internal team members with experience in robotics and autonomous
driving, and external participants with no prior exposure to this
work.  The two participant types provided different and
complementary feedback: internal participants were best able to
evaluate the algorithm itself, whereas external participants offered
first-impression perspectives.  Piloting included 37 participants,
8 of whom were external.
 
The main design decisions that emerged during design for the racing
domain included a trajectory line that changed color depending on
whether the policy inferred the user's intent to pass or stay
behind.  Audio cues were found to overload participants, and it was
difficult for drivers to interpret information about the system's
inferred intent while actively driving.
 
In the urban domain, we built on findings from the racing domain,
using a similar trajectory line and haptic configuration.  A key
focus was the practice and priming setup: we separated onboarding
into three phases (haptic-only, trajectory-only, and combined),
with verbal comprehension checks before advancing to the next phase.
After implementing this change, participant feedback became more
targeted to the policy's actual behavior rather than to the driving
scenario or vehicle physics.  Other key decisions included applying
the safety filter to the trajectory visualization (to prevent
displaying out-of-bounds paths) and scaling the line length by the
magnitude of the assistance.
 
\begin{figure}[t]
    \centering
    \includegraphics[width=0.99\linewidth]{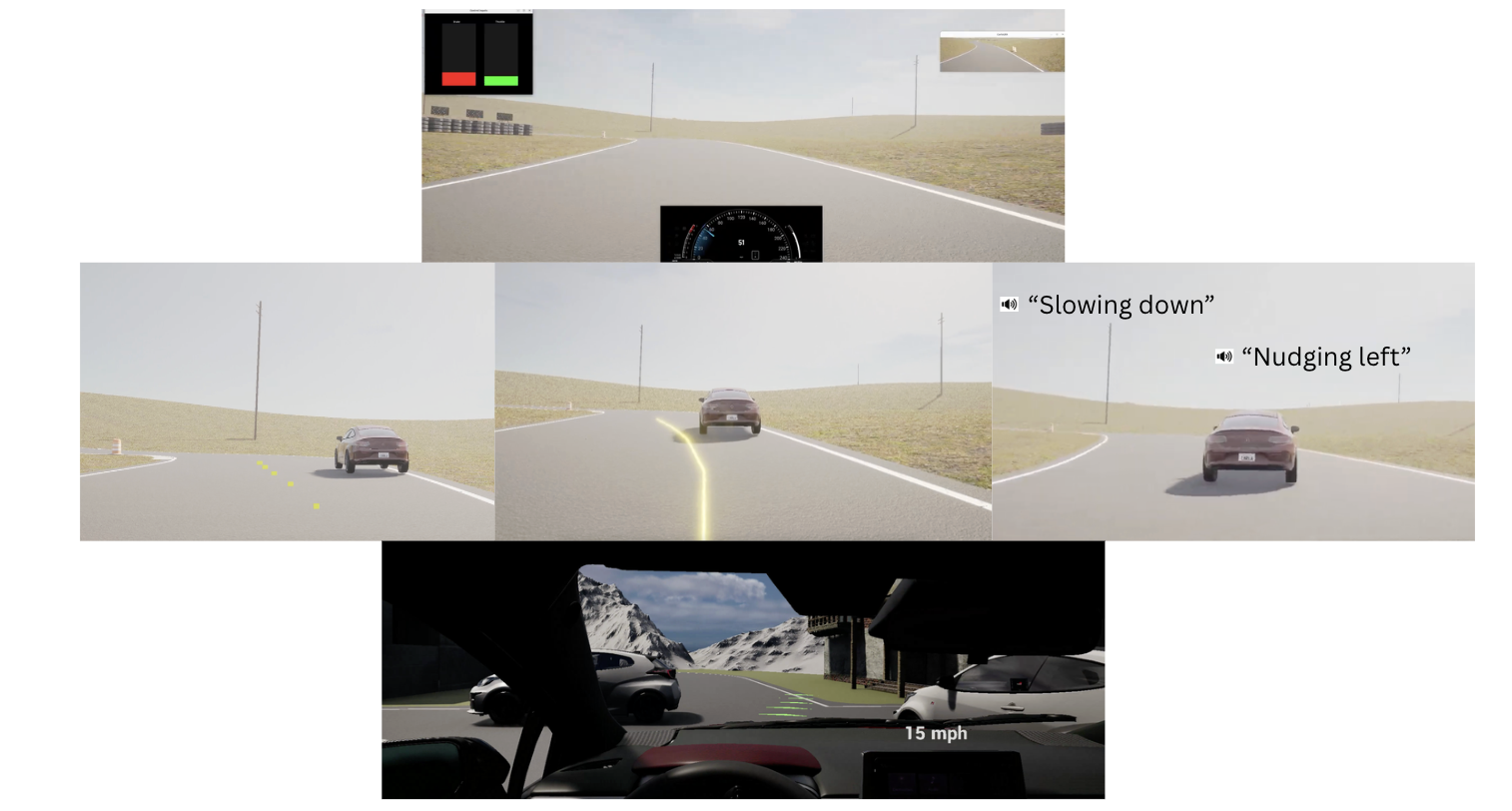}
    \caption{Interfaces tested during iterative piloting for the
      racing domain, showing variations in trajectory line color,
      length, and style.}
    \label{fig:racing_interfaces}
\end{figure}
 
\begin{figure}[t]
    \centering
    \includegraphics[width=0.99\linewidth]{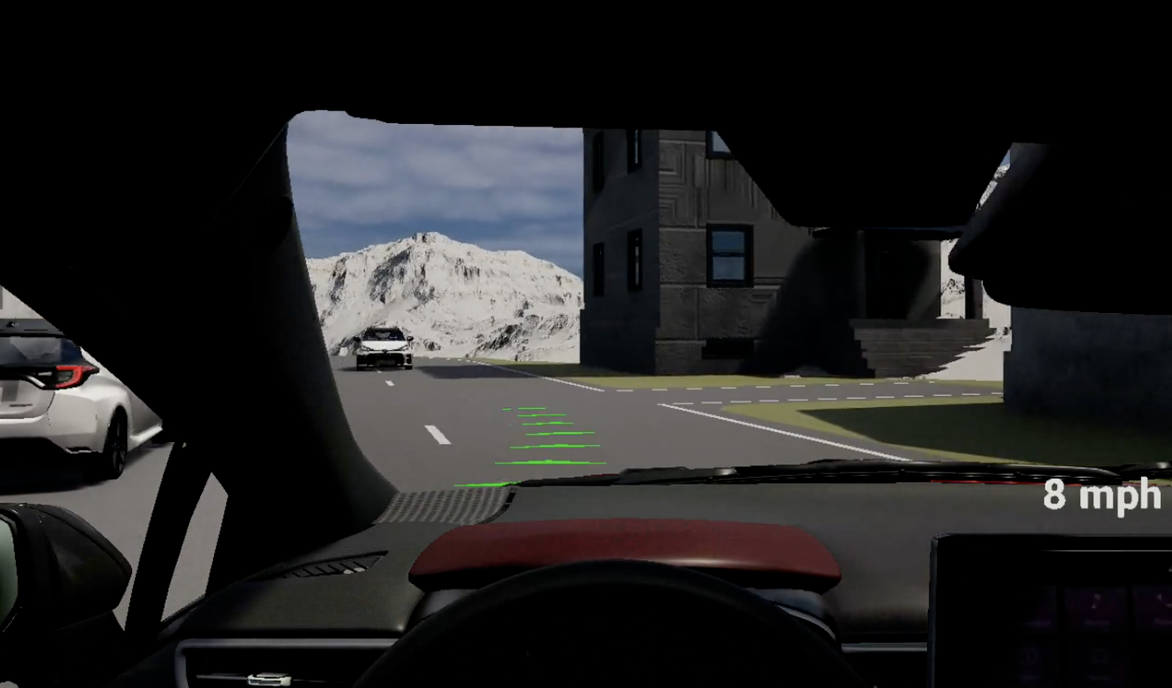}
    \caption{Final trajectory visualization used in Eval~3 (Urban
      Validation).  Line length scales with assistance magnitude;
      green indicates acceleration, red indicates deceleration.}
    \label{fig:urban_final_trajectory}
\end{figure}

%% file: C_detailed_methods.tex
\section{Detailed Methods}
\label{sec:detailed_methods}

\subsection{Methods for Evaluation 1: Racing Characterization}

\paragraph*{Procedure.}
Participants filled out a consent form outlining study risks
(eye strain and motion sickness) and entered the simulator room.
Familiarization proceeded in three phases:
(1)~participants watched one observation lap of the racing
environment;
(2)~they drove one warm-up lap followed by five coached laps, during
which a rule-based AI coach provided directional advice for staying
on the racing line~\citep{kaufman2024effects}; and
(3)~they drove five additional uncoached laps.
The coaching phase ensured that participants, many of whom had no
prior sim-racing experience, could maintain basic vehicle control
before encountering the overtaking task.

After familiarization, participants took a five-minute break and
watched a video explaining the overtaking task.  The task was
divided into blocks defined by track sector and difficulty level.
In each block, participants first followed a set of opponent
vehicles for two laps to observe their behavior and practice
driving the sector, then attempted to overtake the opponent between
marked gates for two laps.  Participants completed one block
without the policy, then watched a video explaining how the policy
would use haptics and a trajectory line to assist overtaking.
The remaining blocks used a staircase
approach~\citep{cornsweet1962staircase}, exposing participants to
up to eight driving scenarios of increasing difficulty;
participants advanced to the next scenario only upon successful
completion, so the number of scenarios completed ranged from one
to eight depending on individual skill level.  Each scenario was
completed both with and without the policy.

After each interaction with the policy, participants rated the
following statements on a 7-point Likert scale (1\,=\,strongly
disagree, 7\,=\,strongly agree):
\begin{enumerate}
  \item The AI knew my intent to pass.
  \item The AI helped me perform the desired maneuver.
  \item The AI's action was safe.
  \item I would have been better off passing without the AI.
\end{enumerate}

\noindent An overall 1--10 rating and open-ended qualitative
feedback were collected after the driving session.
 
\begin{figure}[t]
    \centering
    \includegraphics[width=0.99\linewidth]{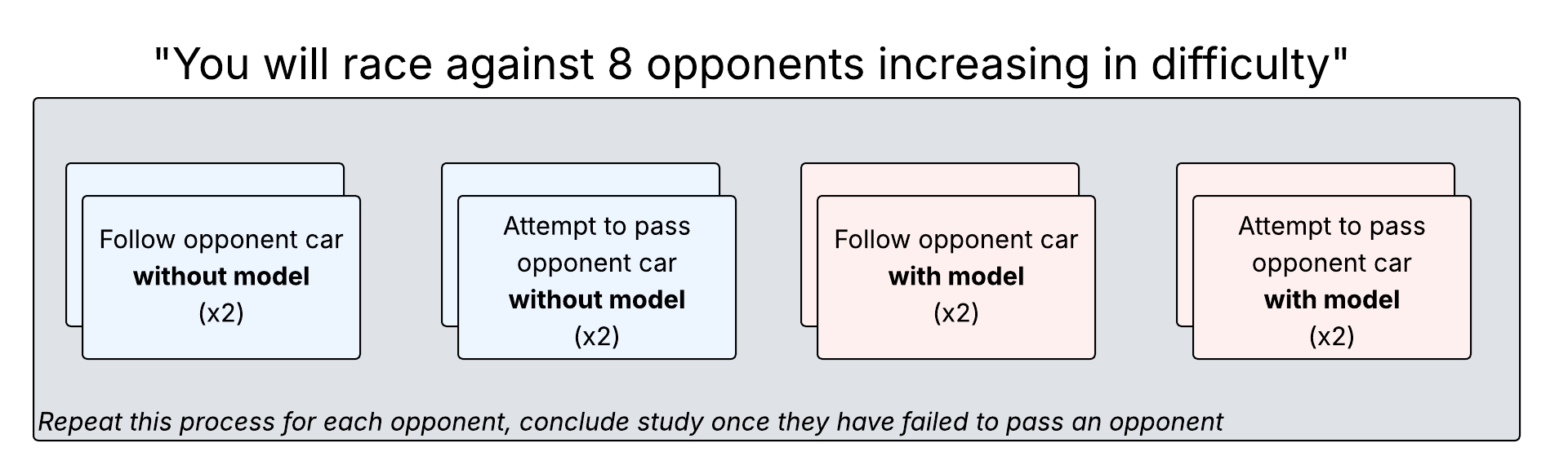}
    \caption{Driving flow for Eval~1 (Racing Characterization).
      Each block consists of two follow laps and two overtake-attempt
      laps, with difficulty increasing via a staircase design.}
    \label{fig:eval1_flow}
\end{figure}

The study flow is illustrated in Fig.~\ref{fig:eval1_flow}.

\paragraph*{Demographics.}
We recruited 33 participants ($M = 45.27$ years; 48.5\% male,
51.5\% female; 51.5\% White, 27.3\% Asian, 12.1\% Black, 9.1\%
Hispanic or Latino).  Two participants did not complete the study
due to motion sickness.  Of the remaining 31, seven were excluded
for not meeting inclusion criteria (completion of all experimental
conditions without simulator sickness, equipment
failure, or not following study guidance).
The final analysis sample was $N = 24$.

\begin{figure}[t]
    \centering
    \includegraphics[width=0.9\linewidth]{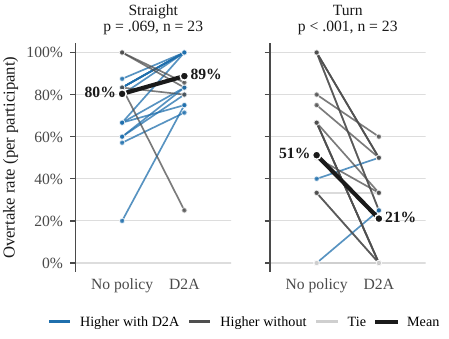}
    \caption{Per-participant overtake rate by track type, without and with the policy, in \textbf{Eval~1} ($n = 23$ participants with both conditions). Thin lines are participants, colored by the direction of change; the thick line is the mean. $p$-values are paired $t$-tests.}
    \label{fig:racing_performance_by_track}
\end{figure}

\begin{figure*}[htbp]
    \centering
    \includegraphics[width=0.99\textwidth, trim=0 0 250 0, clip]{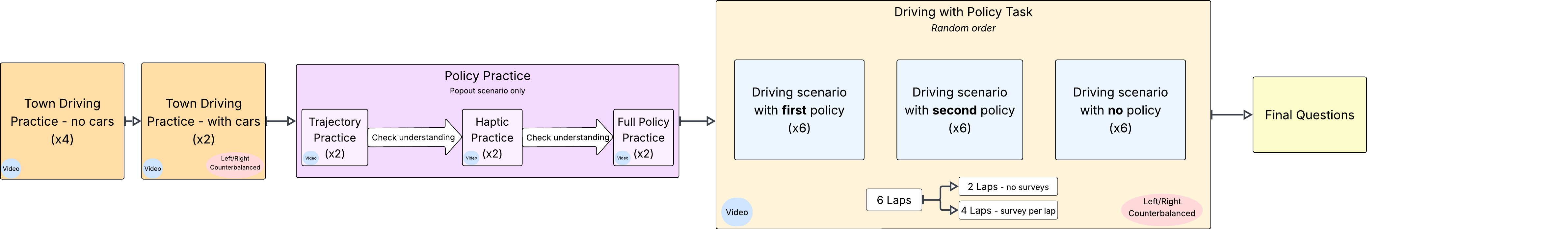}
    \caption{Driving flow for Eval~3 (Urban Validation).}
    \label{fig:eval3_flow}
\end{figure*}

\subsection{Methods for Evaluation 3: Urban Controlled Validation}
\paragraph*{Procedure.}
Participants filled out a consent form outlining study risks (eye
strain and motion sickness) and entered the simulator room.
Familiarization began with a video explaining the town-map rules,
followed by four free-driving laps.  Proctors then introduced
opponent vehicles in two sections of the route: (1)~a car parked on
the right with another vehicle unexpectedly backing out of a
driveway, and (2)~a car parked on the left with an oncoming vehicle
driving the wrong way down the center lane.  Participants drove two
practice laps encountering each scenario twice; proctors ensured
one lap passed the oncoming vehicle on the left and one on the
right.
 
Policy onboarding followed a three-phase protocol.  First,
participants watched a video and drove the driveway scenario twice
with haptic feedback only.  Second, they repeated the procedure
with the trajectory visualization only.  After each phase, proctors
asked participants to describe what the policy was communicating;
if comprehension was insufficient, the behavior was re-explained
and an additional practice lap was given.  Third, participants drove
the scenario twice with both modalities active.
 
For the main task, participants watched a video of the full
scenario and were reminded to compare the policies.  They drove six
laps with each of the three conditions (\withintent{}, \nointent{},
and \nopolicy{}), for a total of 18 laps.  Each condition began
with two laps, followed by a survey, then four additional laps with
a second survey.  Participants alternated passing left and right
around the oncoming vehicle each lap.  Condition order was
randomized in a double-blind fashion.
 
After each set of laps, participants rated the following on a
7-point Likert scale (1\,=\,strongly disagree, 7\,=\,strongly
agree).  Because each lap included multiple opponent-vehicle
encounters (oncoming, driveway pop-out, and roundabout),
Questions~1--4 explicitly instructed participants to consider
\emph{only} the oncoming-car encounter:
 
\begin{enumerate}
  \item Consider ONLY the oncoming car.  The AI assistant was
        providing visual and haptic assistance in the appropriate
        direction I expected.  \emph{(Directionality)}
  \item Consider ONLY the oncoming car.  The AI assistant was
        providing visual and haptic assistance at the appropriate
        time I expected.  \emph{(Timing)}
  \item Consider ONLY the oncoming car.  The AI assistant was
        adapting to which direction I wanted to go.
        \emph{(Adapting Direction)}
  \item Consider ONLY the oncoming car.  I had to fight the AI
        assistant to go in the direction I wanted.
        \emph{(Fight Direction, reverse-coded)}
\end{enumerate}

\noindent Two additional questions addressed overall impressions
across all encounters:
 
\begin{enumerate}[resume]
  \item I felt safe driving with the AI assistant.
        \emph{(Safety)}
  \item I am happy with how well the system worked with me.
        \emph{(Teaming)}
\end{enumerate}
 
\noindent Finally, participants provided an overall policy rating on
a 1--10 scale.

\noindent The study flow is illustrated in Fig.~\ref{fig:eval3_flow}.

\paragraph*{Demographics.}
We recruited 12 participants ($M = 32.80$ years, $SD = 8.64$,
range\,=\,[22--50]; 30.0\% male, 70.0\% female; 60.0\% White,
30.0\% Asian, 10.0\% Black or African American).  One participant
was an internal employee.  Two participants were excluded from the
analysis: one due to motion sickness and one for not following
questionnaire instructions.  The final analysis sample was
$N = 10$.

%% file: D_additional_results.tex
\section{Additional Results}
\label{app:additional_results}
 
This appendix presents supplementary analyses for each evaluation.

\subsection{Eval 1: Racing Characterization}
 
\paragraph{Overtake success by track segment.}
Figure~\ref{fig:racing_performance_by_track} shows mean overtake rates
broken down by track type (straight vs.\ turn) and condition
(\initialpolicy{} vs.\ \nopolicy{}).  On straightaways,
\initialpolicy{} improved overtake success from 80\% to 89\%
($p = .069$); on turns, overtake success decreased from
51\% to 21\% ($p < .001$), confirming the
dynamics-dependent failure mode discussed in
Sec.~\ref{sec:eval1}.

\paragraph{Overtake quality metrics.}
Tables~\ref{tab:overtake_quality} and
\ref{tab:overtake_quality_track} report overtake speed and minimum
opponent clearance.  On straightaways, \initialpolicy{} maintained
comparable overtake speed while increasing clearance
($+0.47$\,m).  On turns, both speed and clearance decreased,
consistent with the policy's tendency to over-steer through turns.

\subsection{Eval 2: Racing Refinement}
 
\begin{figure}
        \begin{subfigure}[b]{0.99\linewidth}
        \includegraphics[width=\linewidth]{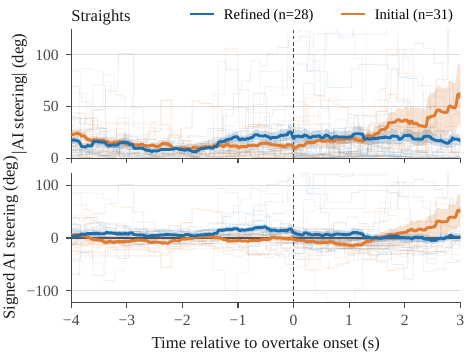}
        \caption{Straights}
        \label{fig:eval_2_torque_traces_straights}
    \end{subfigure}
    
    \begin{subfigure}[b]{0.99\linewidth}
        \includegraphics[width=\linewidth]{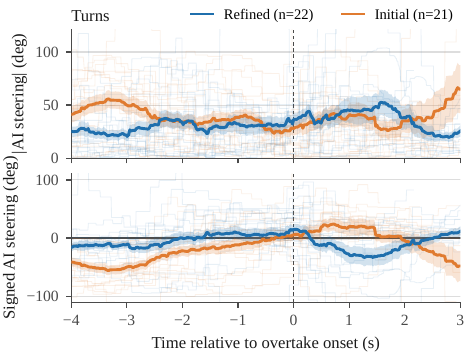}
        \caption{Turns}
        \label{fig:eval_2_torque_traces_turns}
    \end{subfigure}
    \caption{Averaged torque traces across all participants and all (a) \textit{straight} and (b) \textit{turn} trials for \textbf{Eval~2}.  On turns, \initialpolicy{} applies larger steering 4--3\,s before overtake onset ($t = 0$) than \refinedpolicy{}; on straights the two are similar. Traces show raw signed steering, not steering relative to the driver's intent.}
    \label{fig:eval_2_torque_traces_by_segment}
\end{figure}

\paragraph{Torque traces by track segment.}
Figure~\ref{fig:eval_2_torque_traces_by_segment} shows averaged torque traces
for \initialpolicy{} and \refinedpolicy{}, separated by straight
(Fig.~\ref{fig:eval_2_torque_traces_straights}) and turn
(Fig.~\ref{fig:eval_2_torque_traces_turns}) segments.  On turns, \initialpolicy{} applies larger steering than \refinedpolicy{} in the seconds before the overtake ($t < 0$), the regime where the original policy over-steered.  On straightaways, the
two configurations produce similar torque profiles, confirming
that the refinements did not regress baseline performance.

\subsection{Eval 3: Urban Controlled Validation}
 
\begin{figure*}[tbp]
    \centering
    \begin{subfigure}[b]{0.32\linewidth}
        \includegraphics[width=\linewidth]{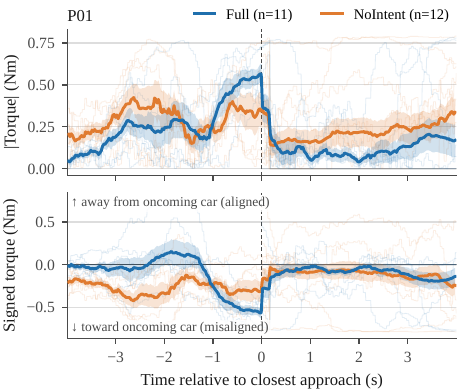}
        \caption{Participant 01}
        \label{fig:eval_3_torque_traces_001}
    \end{subfigure}
    \begin{subfigure}[b]{0.32\linewidth}
        \includegraphics[width=\linewidth]{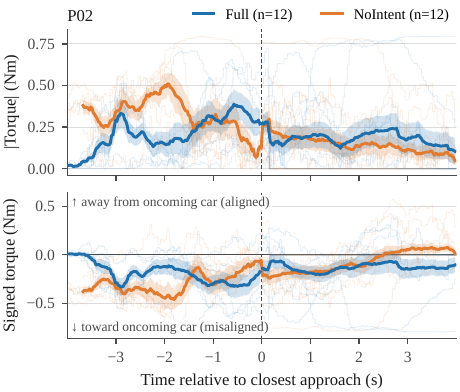}
        \caption{Participant 02}
        \label{fig:eval_3_torque_traces_002}
    \end{subfigure}
    \begin{subfigure}[b]{0.32\linewidth}
        \includegraphics[width=\linewidth]{figures/pid003_ado8_stacked.pdf}
        \caption{Participant 03}
        \label{fig:eval_3_torque_traces_003}
    \end{subfigure}
    \begin{subfigure}[b]{0.32\linewidth}
        \includegraphics[width=\linewidth]{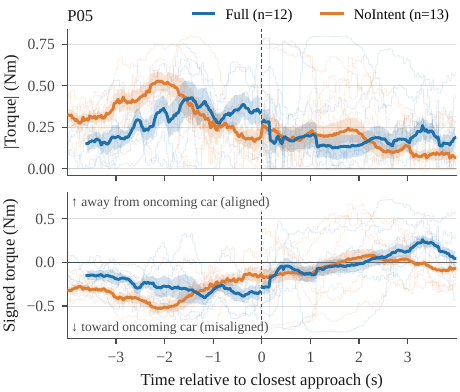}
        \caption{Participant 05}
        \label{fig:eval_3_torque_traces_005}
    \end{subfigure}
    \begin{subfigure}[b]{0.32\linewidth}
        \includegraphics[width=\linewidth]{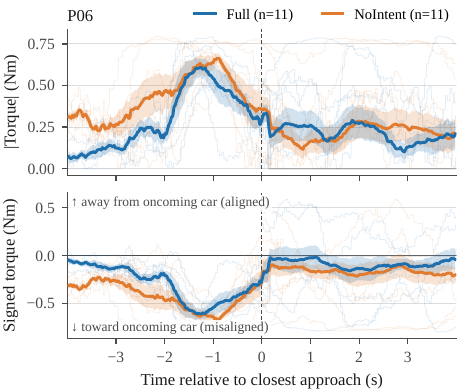}
        \caption{Participant 06}
        \label{fig:eval_3_torque_traces_006}
    \end{subfigure}
    \begin{subfigure}[b]{0.32\linewidth}
        \includegraphics[width=\linewidth]{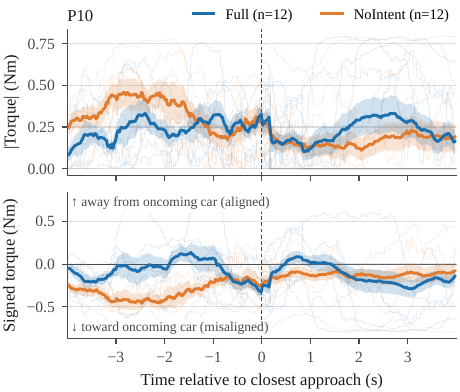}
        \caption{Participant 10}
        \label{fig:eval_3_torque_traces_010}
    \end{subfigure}  
    \begin{subfigure}[b]{0.32\linewidth}
        \includegraphics[width=\linewidth]{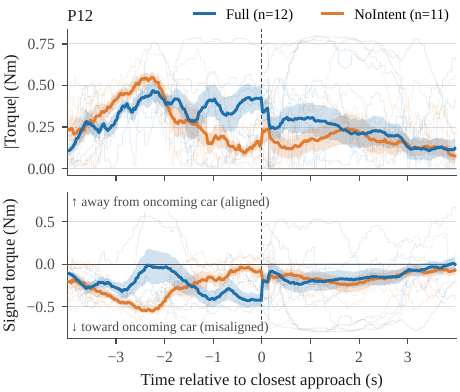}
        \caption{Participant 12}
        \label{fig:eval_3_torque_traces_012}
    \end{subfigure}
    
    \caption{Averaged torque traces for selected participants in the oncoming car event in \textbf{Eval~3}. Top: torque magnitude. Bottom: signed torque, positive away from the oncoming car (aligned with the driver's avoidance). From 3 to 1.5\,s before closest approach, \withintent{} is less misaligned than \nointent{} for all seven participants shown; in the final second, \withintent{} is the more misaligned of the two for five of them (P01, P02, P03, P05, P12).}
    \label{fig:eval_3_torque_traces_per_participant}
\end{figure*}

\paragraph{Per-participant torque traces.}
Figure~\ref{fig:eval_3_torque_traces_per_participant} shows averaged torque traces
for selected participants in the oncoming-vehicle event.  From 3 to 1.5\,s before closest approach, \withintent{} is less misaligned than \nointent{} for all seven participants shown, though in the final second \withintent{} is the more misaligned of the two for five of the participants (Figs.~\ref{fig:eval_3_torque_traces_001}--\ref{fig:eval_3_torque_traces_012}).
\nointent{} steers on average contrary to the driver's
intended avoidance direction (negative signed torque) before the
closest approach ($t = 0$), consistent with the ``captain''
characterization reported in subjective feedback
(Sec.~\ref{sec:eval3_subjective}).  In these plots, PIDs 04, 07, 08, 09, and 11
are excluded due to data quality issues.

\paragraph{Torque magnitude and alignment statistics.}
Figure~\ref{fig:torque_magnitude} shows per-participant torque magnitude
(Fig.~\ref{fig:torque_magnitude_stats}) and
torque-magnitude difference
($\Delta\tau$, Eq.~\eqref{eq:torque_difference};
Fig.~\ref{fig:torque_magnitude_difference}).
Of the 11 participants with both conditions, seven received higher overall torque under \withintent{} and eight experienced more intent-aligned torque, consistent with the aggregate alignment-event counts
reported in Sec.~\ref{sec:eval3}.

\begin{figure}[t]
    \centering
    \includegraphics[width=\columnwidth]{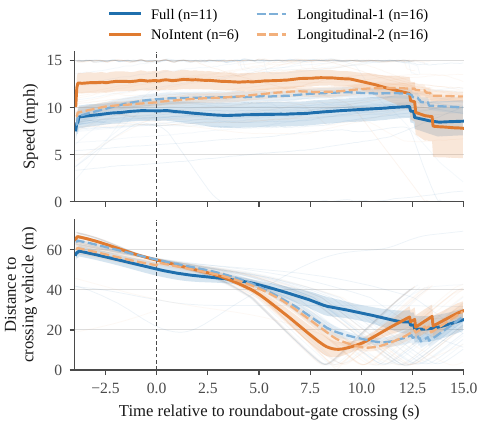}
    \caption{Event-locked traces for the roundabout-approach scenario on GRIP, aligned to the roundabout-gate crossing ($t = 0$).  Top: vehicle speed.  Bottom: distance to the crossing vehicle.  Lines show mean $\pm$ SEM with individual trials dimmed; $n$ = trials per condition.}
    \label{fig:grip_roundabout_traces}
\end{figure}

\subsection{GRIP: Platform Transfer}

\paragraph{Qualitative scenario illustrations.}
Figure~\ref{fig:grip_snapshots}
shows time-ordered snapshots from two representative GRIP trials.
The top row shows lateral steering assistance during an oncoming-vehicle encounter (13–14 mph); the bottom row shows the policy’s braking behavior during a roundabout approach (9 $\rightarrow$ 4 mph in the trial shown).  These behaviors transferred from the desktop
simulator without retraining, confirming that the learned policy
generalizes across platforms with matched dynamics.

\paragraph{Roundabout approach traces.}
Figure~\ref{fig:grip_roundabout_traces} shows event-locked traces
for the roundabout-approach scenario across four GRIP conditions.
\withintent{} and \nointent{} match the conditions
reported in \textbf{Eval~3}.
\textbf{D2A-Longitudinal-1} and \textbf{D2A-Longitudinal-2} are
developmental variants of the intent-conditioned policy that differ
in reward tuning (speed penalties, intervention penalty magnitude,
and collision proximity thresholds); the ``longitudinal'' label
reflects their observed braking behavior rather than a separate
longitudinal intent space.  All intent-conditioned policies use
lateral intents ($\mathcal{Y} = \{\text{left}, \text{right}\}$);
braking emerges from task-level rewards as discussed in
Sec.~\ref{sec:grip_platform}.
The top panel shows vehicle speed.  Drivers' behaviors differed more than the conditions did.  For instance, one participant entered the gate at 12--15\,mph, the others at 8--11\,mph.  In the two sessions that ran both \withintent{} and \nointent{}, \withintent{} crossed the gate 2--3\,mph slower (8.7 vs.\ 10.7 and 12.2 vs.\ 15.0\,mph).  The bottom panel shows that \withintent{} kept more distance to the crossing vehicle on average, though the means mix drivers.  The GRIP logs do not isolate AI-applied braking, so we do not report brake torque.  Given the small sample ($N = 3$), we report these as descriptive observations.